\documentclass[sigconf,nonacm]{acmart}

\AtBeginDocument{%
  \providecommand\BibTeX{{%
    \normalfont B\kern-0.5em{\scshape i\kern-0.25em b}\kern-0.8em\TeX}}}

\usepackage{graphicx}
\usepackage{amsmath}
\usepackage{booktabs}
\usepackage{bigstrut}
\usepackage{xspace}
\usepackage{subcaption}
\usepackage{enumitem}
\usepackage{tabularray}

\usepackage{colortbl}
\definecolor{Gray}{gray}{0.85}
\definecolor{aliceblue}{rgb}{0.94, 0.97, 1.0}
\definecolor{beaublue}{rgb}{0.74, 0.83, 0.9}
\definecolor{blond}{rgb}{0.98, 0.94, 0.75}
\definecolor{beige}{rgb}{0.96, 0.96, 0.86}
\definecolor{cornsilk}{rgb}{1.0, 0.97, 0.86}
\definecolor{platinum}{rgb}{0.9, 0.89, 0.89}
\definecolor{blue(pigment)}{rgb}{0.2, 0.2, 0.6}
\definecolor{goldenbrown}{rgb}{0.6, 0.4, 0.08}

\usepackage{tikz}
\newcommand*\cir[1]{\tikz[baseline=(char.base)]{
            \node[shape=circle,draw,inner sep=1pt] (char) {#1};}}
            
\newcommand{\method}{\texttt{SPOT}\xspace}

\newcommand{\bmX}{\mathcal{X}}
\newcommand{\bmY}{\mathcal{Y}}
\newcommand{\bX}{\mathbf{X}}
\newcommand{\bx}{\mathbf{x}}
\newcommand{\by}{\mathbf{y}}
\newcommand{\bY}{\mathbf{Y}}

\newcommand{\bt}{\mathbf{t}}
\newcommand{\bc}{\mathbf{c}}
\newcommand{\bd}{\mathbf{d}}
\newcommand{\bz}{\mathbf{z}}
\newcommand{\bh}{\mathbf{h}}
\newcommand{\bmH}{\mathcal{H}}
\newcommand{\bH}{\mathbf{H}}

\begin{document}

\title{\method: Sequential Predictive Modeling of Clinical Trial Outcome with Meta-Learning}

\author{Zifeng Wang}
\affiliation{%
\institution{UIUC}
\city{Urbana}
\state{IL}
\country{USA}
}
\email{zifengw2@illinois.edu}

\author{Cao Xiao}
\affiliation{%
\institution{Relativity}
\city{Chicago}
\state{IL}
\country{USA}
}
\email{cao.xiao@relativity.com}

\author{Jimeng Sun}
\affiliation{%
\institution{UIUC}
\city{Urbana}
\state{IL}
\country{USA}
}
\email{jimeng@illinois.edu}

% \affiliation{%

% \affiliation{%

% \affiliation{%

% \affiliation{%

% \affiliation{%

% \affiliation{%

% \affiliation{%

\renewcommand{\shortauthors}{Wang et al.}

\begin{abstract}
Clinical trials are essential to drug development but  time-consuming, costly, and prone to failure. Accurate trial outcome prediction based on historical trial data promises better trial investment decisions and more trial success. Existing trial outcome prediction models were not designed to model the relations among similar trials, capture the progression of features and designs of similar trials, or address the skewness of trial data which causes inferior performance for less common trials. 

To fill the gap and provide accurate trial outcome prediction, we propose \textbf{S}equential \textbf{P}redictive m\textbf{O}deling of clinical \textbf{T}rial outcome (\method) that first identifies trial topics to cluster the multi-sourced trial data into relevant trial topics. It then generates trial embeddings and organizes them by topic and time to create clinical trial sequences. With the consideration of each trial sequence as a task, it uses a meta-learning strategy to achieve a point where the model can rapidly adapt to new tasks with minimal updates. In particular, the topic discovery module enables a deeper understanding of the underlying structure of the data, while sequential learning captures the evolution of trial designs and outcomes. This results in predictions that are not only more accurate but also more interpretable, taking into account the temporal patterns and unique characteristics of each trial topic. We demonstrate that \method wins over the prior methods by a significant margin on trial outcome benchmark data: with a 21.5\% lift on phase I, an 8.9\% lift on phase II, and a 5.5\% lift on phase III trials in the metric of the area under precision-recall curve (PR-AUC). 
\end{abstract}

\maketitle

\section{Introduction}
\label{sec:intro}

Clinical trials are essential for developing new treatments, where drugs must pass the safety and efficacy thresholds across at least three trial phases before they can get approval for manufacturing. Running the trials is time-consuming and costly. On average, it takes about $10\sim15$ years and costs \$2.87 billion for a drug to complete all trial phases, and many drugs fail in clinical trials ~\cite{blass2015basic}. As clinical trial records and their outcomes have been collected and digitized during the past decade, which opens the door for providing more data-driven insights that can ensure more trial success.

\begin{figure}[t]
\centering
\includegraphics[width=\linewidth]{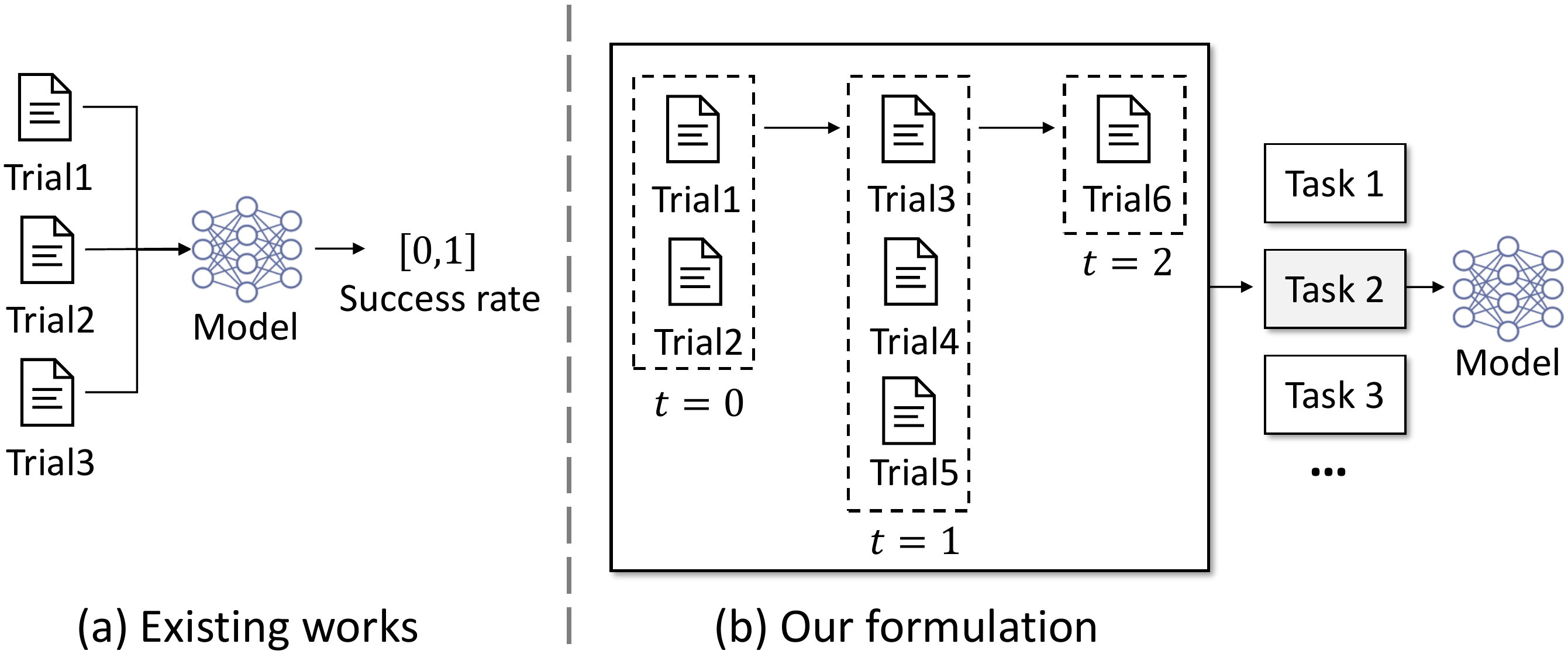}
\caption{Our formulation differs from all existing works: existing works predict trial outcomes for each trial separately; while \method aggregates trials of the same topic to a sequence based on their timestamps, thus utilizing information of similar trials and their progression in predictions. It also considers each trial sequence as a task and mitigates the trial data imbalance problem using a meta-learning strategy. \label{fig:demo_formulation}}
\end{figure}

\begin{figure*}[t]
\centering
\begin{subfigure}[b]{0.3\linewidth}
\includegraphics[width=\linewidth]{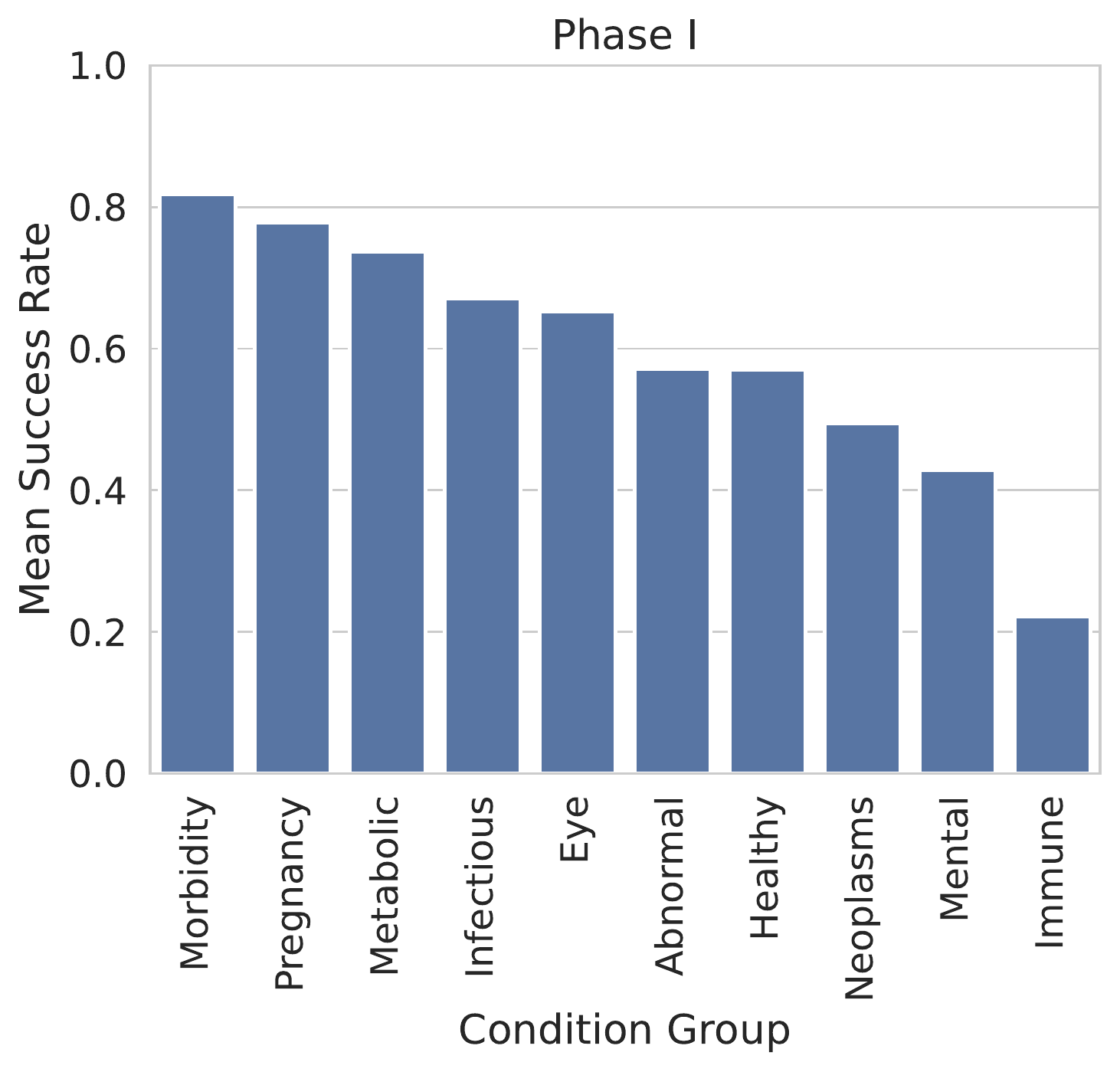}
\end{subfigure}
\hfill
\begin{subfigure}[b]{0.3\linewidth}
\includegraphics[width=\linewidth]{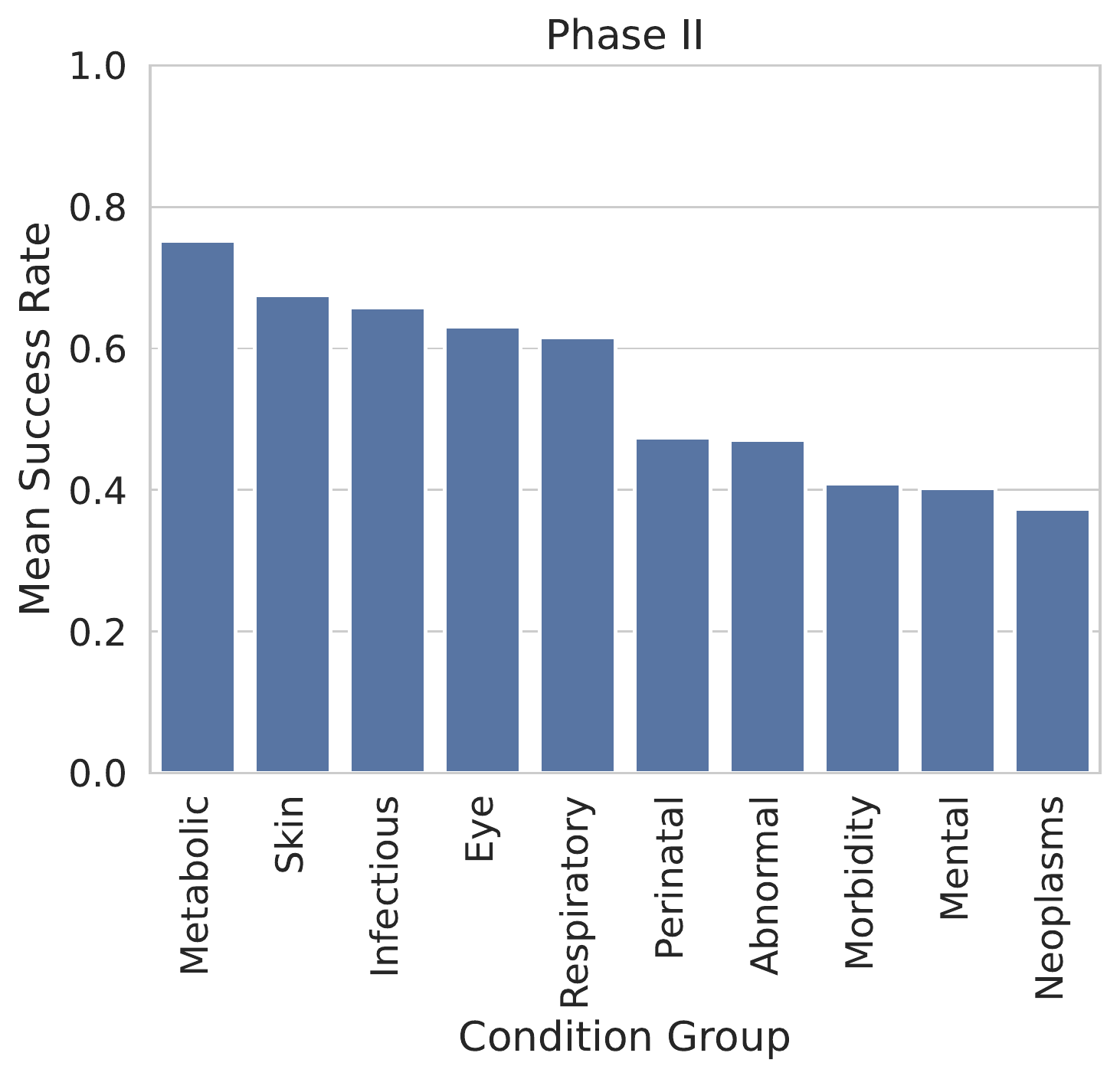}
\end{subfigure}
\hfill
\begin{subfigure}[b]{0.3\linewidth}
\includegraphics[width=\linewidth]{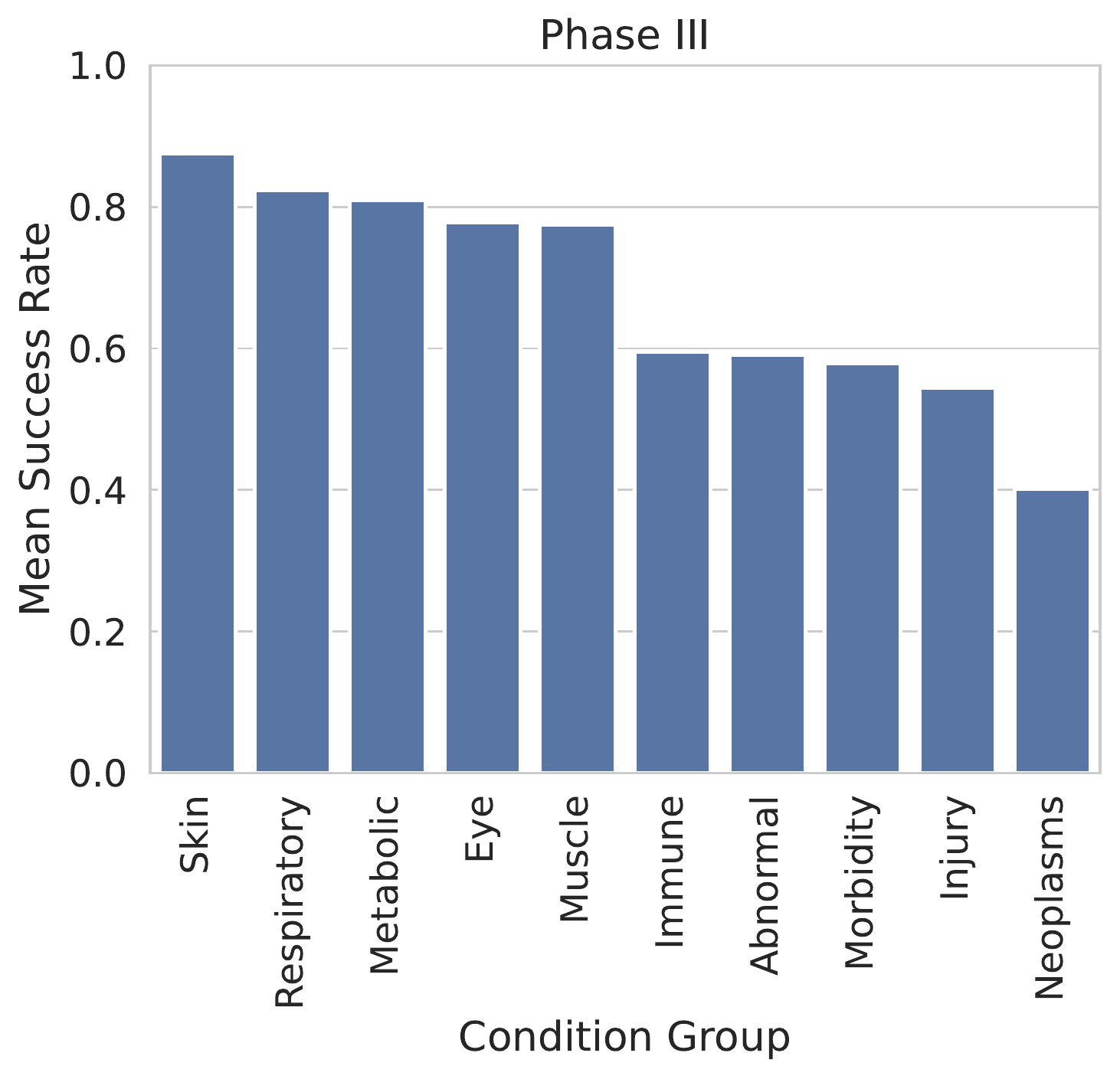}
\end{subfigure}
\caption{
Trial outcome distributions of phase I, II, and III  trials  from the \textit{TOP} dataset \cite{fu2022hint}. We plot target conditions of trials on $x$-axis and the average trial success rates on $y$-axis. It can be seen that trials on different topics have distinct success rates. \label{fig:demo_outcome_dist}
}
\end{figure*}

Among others, clinical trial outcome prediction is one critical task. Earlier attempts were often made to predict individual components in clinical trials to improve the trial results for each trials~\cite{gayvert2016data, qi2019predicting}. Recently, researchers started to propose general methods for trial outcome predictions. For example, ~\citet{Lo2019-eh} predicted drug approvals for 15 disease groups based on drug and clinical trial features using classical machine learning methods. More recently, ~\citet{fu2022hint} proposed leveraging data from multiple sources (e.g., molecule information, trial documents, disease knowledge graph, etc.) with an interaction network to capture their correlations to support trial outcome predictions. Despite these initial successes, they still present the following limitations. 

\begin{itemize}[leftmargin=*]
    \item \textbf{Lack of alleviating heterogeneity in trial patterns}. Clinical trials targeting different diseases and phases could exhibit heterogeneous patterns. A preferred strategy to resolve this heterogeneity is to cluster trials into more homogeneous groups before prediction. Existing works do not consider the grouping. While simply group clinical trials  based on targeting diseases or phases can be inadequate as they may yield too broad categories, where trials under the same branch can vary dramatically in their semantics, or too fine-grained categories, where there are limited numbers of trials in many categories, making it difficult to extract useful patterns.
    \item \textbf{Lack of modeling the progression of trial design for similar trials}. For  similar trials (e.g., targeting disease, drug structure, etc.), the progression of trial designs encodes the knowledge from clinical experts. Modeling this progression pattern could benefit predicting outcomes for new trials. Existing works failed to consider the progression of trial design in their trial outcome prediction. 
    \item \textbf{Lack of handling of the imbalance data }.  Clinical trial data can be highly imbalanced, with specific subgroups or treatments having a small number of trials and diverse average success rates, as illustrated by Fig. \ref{fig:demo_outcome_dist}. This presents machine learning challenge for trial embedding and outcome prediction. Existing methods for outcome prediction struggle to accurately predict outcomes for these minority trials. This is a significant limitation, as it can result in inaccurate predictions and a lack of understanding of the outcomes for these subgroups.
\end{itemize}

To fill the gap and provide accurate trial outcome prediction for all trials, we propose \textbf{S}equential \textbf{P}redictive m\textbf{O}deling of clinical \textbf{T}rial outcome (\method). \method is enabled by three main components:

\begin{itemize}[leftmargin=*]
    \item \textbf{Topic discovery to alleviate  heterogeneous trial patterns.} \method clusters trials into topics where trials of the same topic are more likely to share more homogeneous patterns. This allows for a more nuanced representation of the trials, reducing the noise from dissimilar trials and capturing subtle semantic similarities that pre-defined categories may not capture. 
    \item \textbf{Sequential modeling to capture the progression of trial design.} \method aggregates trials of the same topic to a sequence based on their timestamps and learns to model the temporal patterns of the trial sequence. This allows for extracting knowledge underlying the progression of trial designs and their outcomes, thus enhancing the trial embedding and outcome prediction.
    \item \textbf{Meta learning to model the imbalanced data.}  To alleviate the imbalanced data challenge, \method considers each trial sequence as a task and uses a meta-learning strategy that can generalize well across broad clinical trial distributions.  It allows \method to provide a customized model for each task to maximize the utility. 
\end{itemize}

% The remainder of this paper is organized as follows. We review the literature on AI for clinical trials and meta-learning in \S \ref{sec:related}. We elaborate on the main methods in \S \ref{sec:method}, followed by the experiment results in \S \ref{sec:result}. We end up with a conclusion in \S \ref{sec:conclusion}.

\section{Related Work}\label{sec:related}

\noindent\textbf{Machine Learning for Clinical Trials}.
 Machine learning  has been used in advancing the clinical trial modeling across a diverse set of tasks \cite{wang2022artificial}, including \textit{clinical trial patient-level outcome prediction}~\cite{wang2022transtab}, \textit{clinical trial referential search}~\cite{marshall2017automating,wang2022trial2vec}, and \textit{clinical trial outcome prediction}~\cite{gayvert2016data,siah2021predicting,fu2022hint}. For the task of trial outcome prediction, \citet{gayvert2016data} fused drug chemical structures and properties to predict drug toxicity based on random forests; \citet{qi2019predicting} predicted the pharmacokinetics at phase III using records obtained from phase II trials using RNNs; \citet{siah2021predicting} predicted drug approvals using a combination of drug and trial features; \citet{Lo2019-eh} tried to predict the drug approval with the outcomes of phase II trials or phase III trials using statistical machine learning models like random forests. 
 
Among them, ~\citet{fu2022hint} is most relevant to our work. It aggregates trial data from multiple sources and utilizes a graph neural network on an interaction graph with trial components to fuse multi-modal input data for outcome prediction. However, it does not account for the relations among the trials. Also, the application of average empirical risk minimization urges the model to serve the majority group while neglecting the utility of the minority group. This issue was demonstrated by \citet{fu2022hint} that although the model achieves 0.723 of AUC for all phase III trials on average, it only gets 0.595 of AUC for trials targeting \textit{Neoplasm}. This paper accounts for the imbalanced model performance across trial groups by aggregating trials by topics and leveraging meta-learning. In addition, we consider the progression of trial outcomes within each topic by sequential modeling.\\

\noindent\textbf{Meta-Learning}.
Meta-learning is performed on a set of tasks and leverages prior experiences when tackling a new task. The research on meta-learning could be traced back to decades ago \cite{schmidhuber1987evolutionary,thrun1998learning}. A resurgence of interest in meta-learning raised in recent years with the development of machine learning and deep learning, which includes meta-learning for few-shot learning \cite{ravi2016optimization,gidaris2018dynamic}, long-tailed learning \cite{kang2019decoupling}, reinforcement learning \cite{duan2016rl,gupta2018meta}, neural architecture search \cite{liu2018darts,brock2018smash}, etc. In particular, we consider taking the optimization-based meta-learning \cite{finn2017model,li2017meta,nichol2018first}. One of the most well-known methods is MAML \cite{finn2017model}, which makes iterative two-round updates for local and global parameters, respectively. Through this process, it learns a model that can quickly adapt to new tasks, e.g., the classification of new classes of images, by direct task loss minimization. Our proposed \method utilizes this technique to provide task-specific prediction models for clinical trial outcomes, accounting for the discrepancy of optimal weights across trial tasks. 

\begin{figure*}[t]
\centering
\includegraphics[width=\linewidth]{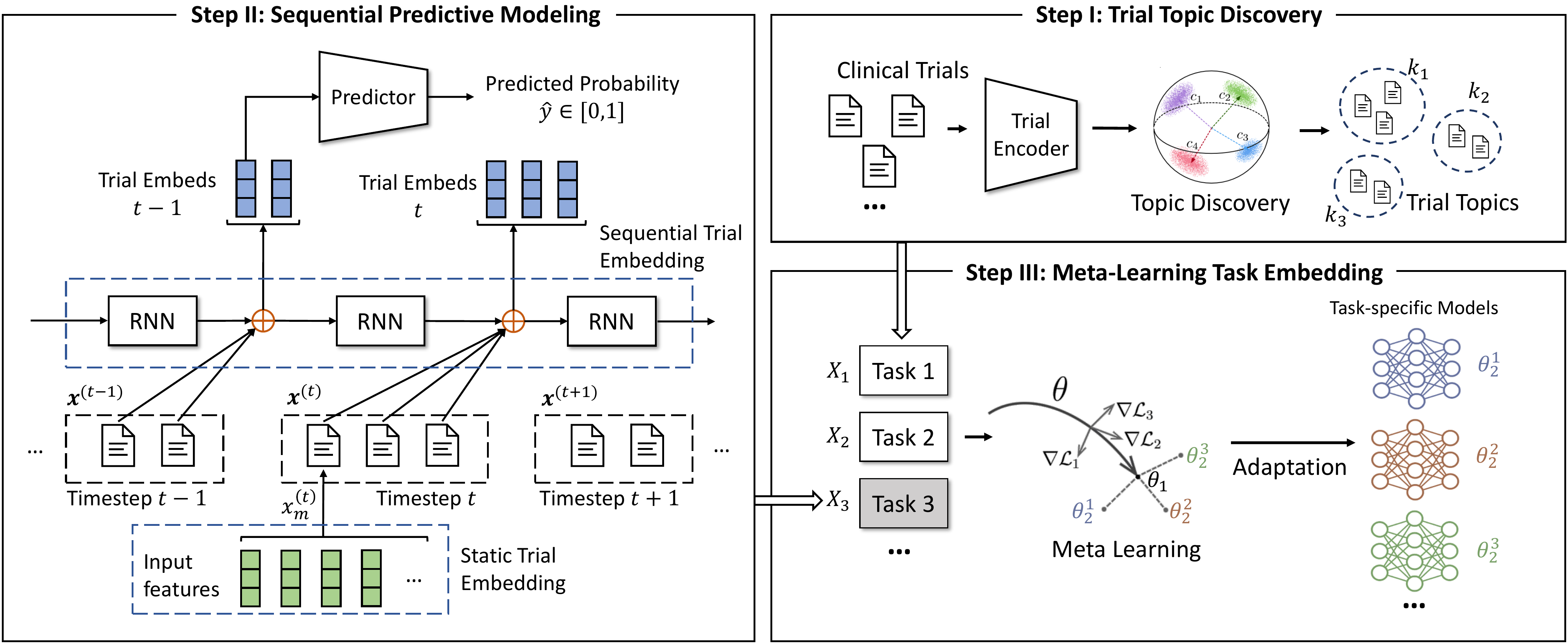}
\caption{\label{fig:method} The illustration of \method workflow. The \method model takes raw clinical trial data with multi-sourced features as input. It first clusters these data into relevant topics using the topic discovery module. The trial embeddings are then generated to prepare data for topic-specific clinical trial sequences. The output trial embeddings are organized based on their topics and arranged chronologically to form clinical trial sequences. We then train a meta-learner based on the sequences, allowing for rapid adaptation to each task with several updates. }
\end{figure*}

\section{The \method Method} \label{sec:method}
The \method model processes raw clinical trial data with multi-sourced features as input then leverages a topic discovery module to cluster the data into relevant topics (Sec \ref{sec:topic_discovery}). It then generates trial embeddings and organizes them by topic and time to create topic-specific clinical trial sequences (Sec \ref{sec:static_trial_emb} and \ref{sec:seq_trial_embed}). Through learning a meta-learner based on these sequences, our model can rapidly adapt to the outcome prediction of incoming sequences of clinical trials with minimal updates (Sec \ref{sec:meta_task_embed}). In the following, we start by presenting the problem formulation and then go with the details of each component one by one.

\subsection{Method Overview} \label{sec:background}
We illustrate the workflow of \method in Fig. \ref{fig:method}. The raw training set $\bmX = \{x_1, \dots, x_N\}$ is a collection of $N$ clinical trials with  corresponding labels $\bmY = \{y_1, \dots, y_N\}$ (i.e., whether a trial reaches its desirable end point). Each trial $x_n$ is represented by three main features as $x_n = \{\bd_n, \bt_n, \bc_n\}$, where $\bd_n = \{d_1,\dots,d_{N_{n}^d}\}$ is a set of target diseases of the $n$-th trial;  $\bt_n = \{t_1,\dots, t_{N_n^t}\}$ is a set of treatments used in the $n$-th trial; $\bc_n = \{c_1, \dots, c_{N_n^c}\}$ is a set of inclusion and exclusion eligibility criteria of the $n$-th trial. $N_{n}^d$, $N_{n}^t$, $N_n^c$ are the total numbers of diseases, treatments, and eligibility criteria in trial $n$, respectively. The following are the two main steps of \method. \\

\noindent \textbf{Topic Discovery}. The first step is to execute topic discovery on $\bmX$ such that we split all trials into $K$ mutually exclusive clusters, as $\bmX = \{\bmX_1,\dots, \bmX_K\}$ and $\bmY = \{\bmY_1,\dots, \bmY_K\}$. For each cluster $\bmX_k$, we rearrange all trials based on their timestamps $t \in \{1,\dots,T_k\}$. Dependening on the time granularity (e.g., by year), there could be multiple concurrent trials with the same timestamp $t$ denoted by the concurrent trial set in topic cluster $k$: $\bx_k^{(t)} = \{x_{k,1}^{(t)},\dots, x_{k,M_{k}^t}^{(t)}\}$, where $M_k^t$ is the total number of trials at the timestamp $t$ in the cluster $k$. We thus obtain a sequence of clinical trials based on $\bmX_k$, which we denote by $\bX_k$ as 
\begin{equation} \label{eq:sequence_topic}
    \bX_k = [\bx_k^{(1)}, \bx_k^{(2)},\dots, \bx_k^{(T_k)}].
\end{equation}
We then model the progression of clinical trial designs and outcomes based on sequences $\bX_k$ for $k = 1,\dots,K$.\\

\noindent \textbf{Trial Outcome Prediction}. The ultimate goal of our method is to predict the outcomes of clinical trials as $\bmY = \{y_1,\dots,y_N\}$, where each $y_n \in \{0,1\}$ indicating the \textit{failure} ($y_n=0$) or \textit{success} ($y_n=1$) of the trial. Here, trial success is defined by whether the trial is completed and meets its primary endpoint.

To predict trial $x_n$ from topic $k$ and timestamp $t$, we first put trial $x_n$ into the proper position in the trial sequence and then predict using the trials before timestamp $t$. Formally we consider $x_n$ be in $\bx_k^{(t)} = \{x_n,x_{k,2}^{(t)},\dots,x_{k,M_k^t}^{(t)}\}$. Therefore, the prediction model $f(\cdot)$ leverages both the features of $x_n$ as $x_n = \{\bd_n, \bt_n, \bc_n\}$ and the precedent trials in the same sequence $\bX_k$, as
\begin{equation}\label{eq:prediction}
    \hat{y}_n = f\left(\{\bd_n, \bt_n, \bc_n\}, \bX_k^{1:t} \right) \in [0,1],
\end{equation}
where $\bX_k^{1:t}$ are all the elements of $\bX_k$ that are before the timestamp $t$, as 
\begin{equation}
    \bX_k^{1:t} = [\bx_k^{(1)},\dots, \bx_k^{(t)}].
\end{equation}
The objective function is hence the binary cross entropy leveraging $y$ as the supervision to $\hat{y}$ for updating $f(\cdot)$.

\subsection{Trial Topic Discovery}\label{sec:topic_discovery}
 As illustrated by Fig. \ref{fig:demo_formulation}, we solicit the latent topic structures in the trial dataset to establish the trial topics, as the trials of the same topic should be similar, considering multiple aspects such as their targeting diseases, treatments, criteria design, etc. We further define the predictive modeling of outcomes for different topics of trials are distinct \textit{tasks}, which comprise a sequence of trials belonging to the same topic.
 
To cluster trials into coherent topics, we utilize a pretrained language model Trial2Vec that generates dense trial embeddings with rich semantics~\cite{wang2022trial2vec}. Formally, we extract the most prominent attributes from the raw trial document, such as the title, disease, treatment, eligibility criteria, and outcome measure, as $x^{\text{attr}}$; and the context information such as the trial description, as $x^{\text{ctx}}$. We first encode all elements in $x^{\text{attr}}$ to dense embeddings as in Eq.~\eqref{eq:denseemb},
\begin{equation}\label{eq:denseemb}
    \bz^{\text{attr}} = \{\bz^{\text{title}}, \bz^{\text{disease}}, \bz^{\text{treatment}}, \bz^{\text{criteria}}, \bz^{\text{outcome}} \}.
\end{equation}
We concatenate all elements in the context $x^{\text{ctx}}$ and encode it to $\bz^{\text{ctx}}$. Combining $\bz^{\text{attr}}$ and $\bz^{\text{ctx}}$ using multi-head attention leads to a semantically meaningful embedding $\bz$ for similar trial searching,  as given by Eq.~\eqref{eq:multihead},
\begin{equation}\label{eq:multihead}
    \bz = \text{MultiHeadAttn} (\bz^{\text{ctx}}, \bz^{\text{attr}}; \mathbf{W}),
\end{equation}
where $\mathbf{W}$ indicates the parameters of the attention layer.

We apply k-means clustering to the trial embeddings $\{\bz_1,\dots,\bz_N\}$, hence each trial is assigned a topic label $k \in \{1,\dots,K\}$, where $K$ is the specified number of topics. For each topic $\bmX_k$, the trials are ordered by the starting year, so as to build a sequence of trials, as $\bX_k$ mentioned in Eq. \eqref{eq:sequence_topic}. To evaluate the quality of the clustering results and pick the best $K$, we can leverage internal clustering metrics such as Sum of Squared Distance and Average Silhouette Coefficient.

% \cx{how to decide on the number of topics? Is there any metric that evaluates the quality of clusters? }

\subsection{Static Trial Embedding} \label{sec:static_trial_emb}

While learning trial topics, we also prepare clinical trial data for trial sequences by first transforming them into static trial embedding. 
Specifically, given a trial $x_n = \{\bd_n, \bt_n, \bc_n\}$ which has three main components: disease, treatment, and criteria, we apply three different encoders to each of these components. 

\noindent\textbf{Disease Embedding.} The disease set $\bd_n = \{d_1,\dots,d_{N_n^d}\}$ contains the ICD codes that can be mapped back to a hierarchy of conceptions. We adopt a graph-based attention model named GRAM \cite{choi2017gram} to obtain the code embeddings considering their positions in the knowledge graph as in Eq.~\eqref{eq:kg},
\begin{equation}\label{eq:kg}
    \bh_n^d = \frac1{N_n^d} \sum_{i}^{N_n^d} \text{GRAM}(d_i) \in \mathbb{R}^p,
\end{equation}
where we take a mean-pooling of all code embeddings to get the disease embedding $\bh_n^d$. 

\noindent \textbf{Treatment Embedding.} We adopt a graph message passing neural network (MPNN) \cite{gilmer2017neural} to encode the treatment set that consists of drug molecules represented by SMILES strings, then also take a mean-pooling to obtain the treatment embedding in Eq.~\eqref{eq:treatment},
\begin{equation}\label{eq:treatment}
    \bh_n^t = \frac1{N_n^t} \sum_{i}^{N_n^t} \text{MPNN}(t_i) \in \mathbb{R}^p.
\end{equation}
In this work, we focus on small molecule treatments and plan to explore other treatment embedding for other treatment types such as biologics and medical devices as future work. 

\noindent \textbf{Criteria Embedding.} We leverage the pretrained Trial2Vec \cite{wang2022trial2vec} to encode criteria text $\bc_n=[c_1,\dots,c_{N_n^c}]$ as shown in Eq.~\eqref{eq:criteria},
\begin{equation}\label{eq:criteria}
    \bh_n^c = \frac1{N_n^c} \sum_{i}^{N_n^c} \text{Trial2Vec}(c_i) \in \mathbb{R}^p,
\end{equation}
which we also take mean-pooling.

We also encode the trial's starting year by $\bh_y = \text{year}* \bh_{ts}$, where $\bh_{ts}$ is a trainable numerical embedding. The year embedding $\bh_y$ acts as a representation of the trend of trial outcomes and advances over time because scientific advancements can vary from year to year. Thus, the trial embedding is produced by a highway neural network \cite{srivastava2015highway} that adaptively adjusts the importance of different components through gating as in Eq.~\eqref{eq:trial_embed},
\begin{equation} \label{eq:trial_embed}
    \bh_n = \frac14 \sum \text{Highway}([\bh_n^d \oplus \bh_n^t \oplus \bh_n^c \oplus \bh_{n}^y]) \in \mathbb{R}^p.
\end{equation}
Here, $\oplus$ indicates the concatenation of the input embeddings; we take an average pooling of the concatenated embeddings which are with a shape of $4\times p$ to obtain an overall trial representation $\bh_n \in \mathbb{R}^p$.

\subsection{Sequential Trial Embedding} 
\label{sec:seq_trial_embed}
With the trial topics and the static trial embedding, we are now ready to construct the sequential trial embeddings.
As illustrated by the left part of Fig. \ref{fig:method}, a collection of clinical trials belonging to the same topic are placed in the temporal order, as $\bX_k = [\bx_k^{(1)},\dots, \bx_k^{(T_k)}]$. At timestep $t$, $\bx_{k}^{(t)} = \{x_{k,1}^{(t)},\dots,x_{k,M_k^t}^{(t)} \}$ is a set of trials occurred concurrently. From Eq. \eqref{eq:trial_embed}, we encode trials in $\bx_{k}^{(t)}$ to get the static embeddings as $\bH^{(t)}_k = [\bh^{(t)}_{k,1},\dots, \bh^{(t)}_{k,M_k^t}]$ for all $t \in \{1,\dots,T\}$. 

As the sequential embedding was calculated within a topic, we omit the cluster id $k$ to avoid clutter from now on. The static encoding of a topic $\bX$ can be written by $\bmH=[\bH^{(1)},\dots,\bH^{(T)}]$. We propagate the historical trial information along the timesteps through a recurrent neural network (RNN) as given by Eq.~\eqref{eq:propogate}, 
\begin{equation}\label{eq:propogate}
    \bh^{(t)} = \text{RNN}(\bh^{(t-1)}, \dots, \bh^{(1)}),
\end{equation}
where $\bh^{(t)}$ is initialized by the mean-pooling of all trial embeddings $\bH^{(t)}$ at the timestamp.

In this way, $\bh^{(t)}$ encodes the temporal trial information and can contribute to a spatial interaction with the static embeddings $\bh$ obtained by Eq. \eqref{eq:trial_embed}, as 
\begin{equation}\label{eq:spatial}
    [\widetilde{\bh}^{(t)} \oplus \widetilde{\bH}^{(t)}] = \text{Interaction}([\bh^{(t)} \oplus \bH^{(t)}]) \in \mathbb{R}^{(M^t+1) \times d}.
\end{equation}
Here, $\text{Interaction}(\cdot)$ is a multi-head attention layer that dynamically fuses the temporal information to different trial embeddings $\widetilde{\bH}^{(t)}$. 

We predict trial outcomes via another function $\text{NN}_{\text{pred}}(\cdot)$ that is a one-layer fully-connected neural network:
\begin{equation}\label{eq:pred}
    \hat{\by}^{(t)} = \text{NN}_{\text{pred}}(\widetilde{\bH}^{(t)}) = [\hat{y}_1^{(t)}, \dots ,\hat{y}_{M^t}^{(t)}] \in [0,1]^{M^t},
\end{equation}
which takes the concatenation of the adjusted embeddings and the static embeddings. $M$ is the number of trials of the topic. Applying the prediction model to the whole topic $\bX$ yields predictions as
\begin{equation}
    \hat{\bY} = [\hat{\by}^{(1)}, \dots, \hat{\by}^{(T)}],
\end{equation}
which corresponds to trials $\bX=[\bx^{(1)}, \dots, \bx^{(T)}]$.

\subsection{Meta-Learning Task Embedding}
\label{sec:meta_task_embed}
Inspired by the concept of meta-learning and the distinction across trial topics, we put model-agnostic meta-learning (MAML) \cite{finn2017model} to the \method framework to learn task-specific models. Concretely, we assume that modeling on each trial topic is a separate \textit{task} with two sets of model parameters: global parameters $\{\theta_1,\theta_2\}$ and task-specific parameters $[\theta_2^{1},\dots,\theta_2^{K}]$. All tasks share the same $\theta_1$ was determined in the static trial embedding modules described in \S \ref{sec:static_trial_emb}, e.g., the disease, treatment, and criteria encoder. $\theta_2$ corresponds to the parameters in the RNN and sequential prediction modules. To reflect the progression patterns of topics, \method updates $\theta_2^{k}$ when adapted to the $k$-th task. Formally, the algorithm first samples a batch of tasks $\mathcal{X}$, then it undergoes two steps:\\

\noindent \textbf{Step \cir{1}:} Conduct local updates for each task $\bX_k \sim \mathcal{X}$ via stochastic gradient descent as given by Eq.~\eqref{eq:stocupdate},
\begin{equation}\label{eq:stocupdate}
    \theta_2^k \gets \theta_2^k - \alpha \nabla_{\theta_2^k} \mathcal{L}(\hat{\bY}_k, \bY_k),
\end{equation}
where $\theta_2^k$ is initialized using $\theta_2$; $\hat{\bY}_k$ is predicted by the task-specific model $f_{\theta_1,\theta_2^k}$; the loss $\mathcal{L}(\cdot)$ is binary cross entropy  given by Eq.~\eqref{eq:crossentropyloss},
\begin{equation}\label{eq:crossentropyloss}
    \mathcal{L}(\hat{\bY}, \bY) = -\bY \log \hat{\bY} - (1- \bY) \log (1- \hat{\bY}),
\end{equation}
where $\bY \in \{0, 1\}$: $\bY=0$ trial failed and $\bY=1$  trial succeeded.\\

\noindent \textbf{Step \cir{2}:} Conduct global updates of $\theta_1$ and $\theta_2$ based on the task-specific parameters as shown in Eq.~\eqref{eq:globalupdate},
\begin{equation}\label{eq:globalupdate}
\begin{aligned}
    \theta_1 &\gets \theta_1 - \beta \sum_{\bX_k \in \mathcal{X}} \nabla_{\theta_1} \mathcal{L}(\hat{\bY}_k, \bY), \\
    \theta_2 &\gets \theta_2 - \beta \sum_{\bX_k \in \mathcal{X}} \nabla_{\theta_2} \mathcal{L}(\hat{\bY}_k, \bY). 
\end{aligned}
\end{equation}
Here, $\hat{\bY}_k$ is predicted by $f_{\theta_1,\theta_2^k}$. We take an average of all task-specific parameter gradients to update the global parameter $\theta_2$.

The above steps repeat until both $\theta_1$ and $\theta_2$ converges. Therefore, we can freeze $\theta_1$ and start from $\theta_2$ to quickly adapt the model to every task to obtain the task-specific model. It is based on the intuition that the basic trial property representations should keep stable while the topic style may vary. This method also enjoys fast adaption to a new task with a small query set and a few updates.

\section{Experiment} \label{sec:result}


\subsection{Experimental Setup}

\noindent\textbf{Dataset.}
We utilize the TOP clinical trial outcome prediction benchmark from \citet{fu2022hint}. The dataset includes information on drug, disease, and eligibility criteria for a total of 17,538 clinical trials. The trials are divided into three phases: Phase I (1,787 trials), Phase II (6,102 trials), and Phase III (4,576 trials). Success rates vary by phase, with 56.3\% of phase I trials, 49.8\% of phase II trials, and 67.8\% of phase III trials resulting in success. The distribution of the targeting diseases is available at Table \ref{appx:tab:disease_group_stats}. We conduct experiments on different phases of trials separately.\\

\begin{table}[h!]
  \centering
  \caption{Dataset statistics. \# is short for the number of. \# Trials are shown by the split of train/validation/test sets.}
       \resizebox{\linewidth}{!}{%
    \begin{tabular}{lccccc}
    \toprule
          & \#  \textbf{Trials} & \# \textbf{Drugs} & \# \textbf{Diseases} & \# \textbf{Success} & \# \textbf{Failure} \bigstrut\\
    \midrule
    Phase I & 1,044/116/627 & 2,020 & 1,392 & 1,006 & 781 \bigstrut[t]\\
    Phase II & 4,004/445/1,653 & 5,610 & 2,824 & 3,039 & 3,063 \\
    Phase III & 3,092/344/1,140 & 4,727 & 1,619 & 3,104 & 1,472 \bigstrut[b]\\
    \bottomrule
    \end{tabular}%
    }
  \label{tab:data_stats}%
\end{table}%

\noindent\textbf{Metrics.}
We consider the following performance metrics.
\begin{itemize}[leftmargin=*]
    \item \textbf{AUROC}: the area under the Receiver Operating Characteristic curve. It summarizes the trade-off between the true positive rate (TPR) and the false positive rate (FPR) with the varying threshold of FPR. In theory, it is equivalent to calculating the ranking quality by the model predictions to identify the true positive samples. However, better AUROC does not necessarily indicate better outputting of well-calibrated probability predictions.
    \item \textbf{PRAUC}: the area under the Precision-Recall curve. It summarizes the trade-off between the precision (PPV) and the recall (TPR) with the varying threshold of recall. It is equivalent to the average of precision scores calculated for each recall threshold and is more sensitive to the detection quality of true positives from the data, e.g., identifying which trial is going to succeed.
    \item \textbf{F1-score}: the harmonic mean of precision and recall. It cares if the predicted probabilities are well-calibrated as exact as PRAUC.
\end{itemize}


\noindent\textbf{Baseline.}
We compare \method against the following machine learning and deep learning models, mostly following the setups in \cite{fu2022hint}.

\begin{itemize}[leftmargin=*]
    \item LR \cite{Lo2019-eh}: logistic regression with the default hyperparameters implemented by scikit-learn \cite{scikit-learn}.
    \item RF \cite{Lo2019-eh}: random forests implemented by scikit-learn \cite{scikit-learn}.
    \item XGBoost \cite{siah2021predicting}: a gradient-boosted decision tree method.
    \item AdaBoost \cite{fan2020application}: an adaptive boosting-based decision tree method implemented by scikit-learn \cite{scikit-learn}.
    \item kNN+RF \cite{Lo2019-eh}: handling missing data by imputation with the k nearest neighbors and predict using random forests.
    \item FFNN \cite{tranchevent2019deep}: a feedforward neural network that uses the same features as in~\cite{fu2022hint}. It has 3 fully-connected layers with the dimensions of dim-of-input-feature, 500, and 100, and ReLU activations.
    \item DeepEnroll \cite{zhang2020deepenroll}: it was adapted here for encoding clinical trial eligibility criteria: a hierarchical embedding network to encode disease ontology, and an alignment model to capture interactions between eligibility criteria and disease information. We concatenate the criteria embedding with the molecule embedding obtained by  MPNN for adapting it to trial outcome prediction.
    \item COMPOSE \cite{gao2020compose}: it was originally used for patient-trial matching based on a convolutional neural network and a memory network to encode eligibility criteria and diseases, respectively. We also concatenate its embedding with a molecule embedding by MPNN for trial outcome prediction.
    \item HINT \cite{fu2022hint}: it is the state-of-the-art trial outcome prediction model that incorporates (1) a drug molecule encoder based on MPNN; (2) a disease ontology encoder based on GRAM; (3) a trial eligibility criteria encoder based on BERT; (4) a drug molecule pharmacokinetic encoder; (5) graph neural network for feature interactions. Then, it feeds the interacted embeddings to a prediction model for outcome predictions.
\end{itemize}

To summarize, embeddings from HINT are used in all statistical machine learning models (LR, RF, XGBoost, ADAboost, and kNN+RF) as the input for predicting trial outcomes, while the deep learning models are added with an additional molecule encoder.

\subsection{Results}

We conduct experiments to evaluate \method in the following aspects:

\begin{itemize}[leftmargin=*]
    \item \textbf{EXP 1.} The overall performance of outcome predictions for phase I, II, and III trials, respectively, compared with other baselines.
    \item \textbf{EXP 2.} Analysis of the disparity of prediction performances across different trial groups.
    \item \textbf{EXP 3.} The in-depth analysis of \method in terms of its sensitivity and its sequential modeling ability.
    \item \textbf{EXP 4.} The ablation of the different components of \method.
\end{itemize}

\subsection*{Exp 1. Performance of Trial Outcome Predictions} \label{sec:exp_overall}

We show the trial outcome prediction performances of all compared methods in Tables \ref{tab:result_phase_I}, \ref{tab:result_phase_II}, \ref{tab:result_phase_III}, for phase I, II, and III trials, respectively. The results observed show that \method, demonstrates superior performance over all the baselines across all metrics. Specifically, \method yields a significant performance lift among phase I trials over baselines. In particular, preventing a failing phase I trial can have a substantial impact on the entire drug development process. Accurate prediction of Phase I trial outcomes can allow us to identify more promising clinical trials and improve the chance of discovering new drugs.

\begin{table}[t]
  \centering
  \caption{Trial outcome prediction results compared between \method and baselines for \textbf{phase I trials}.}
     \resizebox{\linewidth}{!}{%
    \begin{tabular}{lccc}
    \toprule
    \multicolumn{4}{c}{ \textbf{Phase I Trials}} \bigstrut[t]\\
    \multicolumn{4}{l}{\# train: 1,044; \# valid: 116; \# test: 627; \# patients per trial: 45} \bigstrut[b]\\
    \midrule
    \textbf{Method} & \textbf{PR-AUC} & \textbf{F1}    & \textbf{ROC-AUC} \bigstrut[t]\\
    LR    & 0.500 ± 0.005 & 0.604 ± 0.005 & 0.520 ± 0.006 \\
    RF    & 0.518 ± 0.005 & 0.621 ± 0.005 & 0.525 ± 0.006 \\
    XGBoost & 0.513 ± 0.06 & 0.621 ± 0.007 & 0.518 ± 0.006 \\
    AdaBoost & 0.519 ± 0.005 & 0.622 ± 0.007 & 0.526 ± 0.006 \\
    kNN+RF & 0.531 ± 0.006 & 0.625 ± 0.007 & 0.538 ± 0.005 \\
    FFNN  & 0.547 ± 0.010 & 0.634 ± 0.015 & 0.550 ± 0.010 \\
    DeepEnroll & 0.568 ± 0.007 & 0.648 ± 0.011 & 0.575 ± 0.013 \\
    COMPOSE & 0.564 ± 0.007 & 0.658 ± 0.009 & 0.571 ± 0.011 \\
    HINT  & 0.567 ± 0.010 & 0.665 ± 0.010 & 0.576 ± 0.008 \\
    \rowcolor[rgb]{ .906,  .902,  .902} SPOT  & \textbf{0.689 ± 0.009} & \textbf{0.714 ± 0.011} & \textbf{0.660 ± 0.008} \bigstrut[b]\\
    \bottomrule
    \end{tabular}%
    }
  \label{tab:result_phase_I}%
\end{table}%

\begin{table}[t]
  \centering
  \caption{Trial outcome prediction results compared between \method and baselines for \textbf{phase II trials}.}
       \resizebox{\linewidth}{!}{%
    \begin{tabular}{lccc}
    \toprule
    \multicolumn{4}{c}{\textbf{Phase II Trials}} \bigstrut[t]\\
    \multicolumn{4}{l}{\# train: 4,004; \# valid: 445; \# test: 1,653; \# patients per trial: 183} \bigstrut[b]\\
    \midrule
   \textbf{Method} & \textbf{PR-AUC} & \textbf{F1}    & \textbf{ROC-AUC} \bigstrut[t]\\
    LR    & 0.565 ± 0.005 & 0.555 ± 0.006 & 0.587 ± 0.009 \\
    RF    & 0.578 ± 0.008 & 0.563 ± 0.009 & 0.588 ± 0.009 \\
    XGBoost & 0.586 ± 0.006 & 0.570 ± 0.009 & 0.600 ± 0.007 \\
    AdaBoost & 0.586 ± 0.009 & 0.583 ± 0.008 & 0.603 ± 0.007 \\
    kNN+RF & 0.594 ± 0.008 & 0.590 ± 0.006 & 0.597 ± 0.008 \\
    FFNN  & 0.604 ± 0.010 & 0.599 ± 0.012 & 0.611 ± 0.011 \\
    DeepEnroll & 0.600 ± 0.010 & 0.598 ± 0.007 & 0.625 ± 0.008 \\
    COMPOSE & 0.604 ± 0.007 & 0.597 ± 0.006 & 0.628 ± 0.009 \\
    HINT  & 0.629 ± 0.009 & 0.620 ± 0.008 & \textbf{0.645 ± 0.006} \\
    \rowcolor[rgb]{ .906,  .902,  .902} SPOT  & \textbf{0.685 ± 0.010} & \textbf{0.656 ± 0.009} & 0.630 ± 0.007 \bigstrut[b]\\
    \bottomrule
    \end{tabular}%
    }
  \label{tab:result_phase_II}%
\end{table}%

\begin{table}[t]
  \centering
  \caption{Trial outcome prediction results compared between \method and baselines for \textbf{phase III trials}.}
       \resizebox{\linewidth}{!}{%
    \begin{tabular}{lccc}
    \toprule
    \multicolumn{4}{c}{\textbf{Phase III Trials}} \bigstrut[t]\\
    \multicolumn{4}{l}{\# train: 3,092; \# valid: 344; \# test: 1,140; \# patients per trial: 1,418} \bigstrut[b]\\
    \midrule
    \textbf{Method} & \textbf{PR-AUC} & \textbf{F1}    & \textbf{ROC-AUC} \bigstrut[t]\\
    LR    & 0.687 ± 0.005 & 0.698 ± 0.005 & 0.650 ± 0.007 \\
    RF    & 0.692 ± 0.004 & 0.686 ± 0.010 & 0.663 ± 0.007 \\
    XGBoost & 0.697 ± 0.007 & 0.696 ± 0.005 & 0.667 ± 0.005 \\
    AdaBoost & 0.701 ± 0.005 & 0.695 ± 0.005 & 0.670 ± 0.004 \\
    kNN+RF & 0.707 ± 0.007 & 0.698 ± 0.008 & 0.678 ± 0.010 \\
    FFNN  & 0.747 ± 0.011 & 0.748 ± 0.009 & 0.681 ± 0.008 \\
    DeepEnroll & 0.777 ± 0.008 & 0.786 ± 0.007 & 0.699 ± 0.008 \\
    COMPOSE & 0.782 ± 0.008 & 0.792 ± 0.007 & 0.700 ± 0.007 \\
    HINT  & 0.811 ± 0.007 & 0.847 ± 0.009 & \textbf{0.723 ± 0.006} \\
    \rowcolor[rgb]{ .906,  .902,  .902} SPOT  & \textbf{0.856 ± 0.008} & \textbf{0.857 ± 0.008} & 0.711 ± 0.005 \bigstrut[b]\\
    \bottomrule
    \end{tabular}%
    }
  \label{tab:result_phase_III}%
\end{table}%

\begin{figure*}[t]
\centering
\begin{subfigure}[b]{0.33\linewidth}
\includegraphics[width=\linewidth]{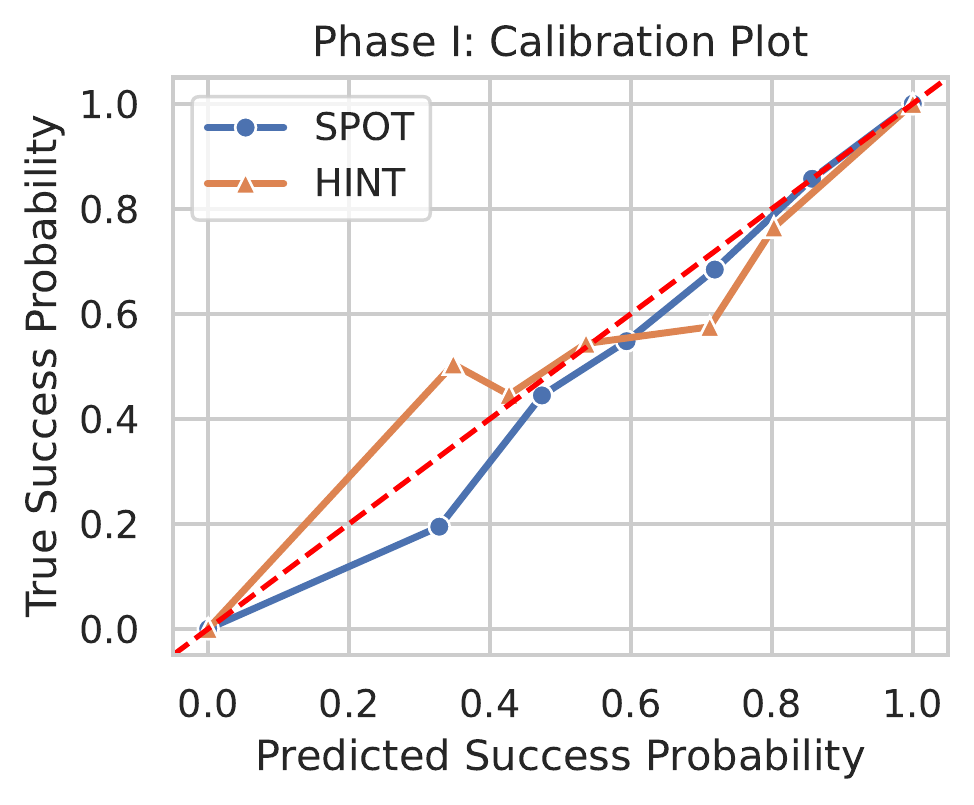}
\end{subfigure}
\hfill
\begin{subfigure}[b]{0.33\linewidth}
\includegraphics[width=\linewidth]{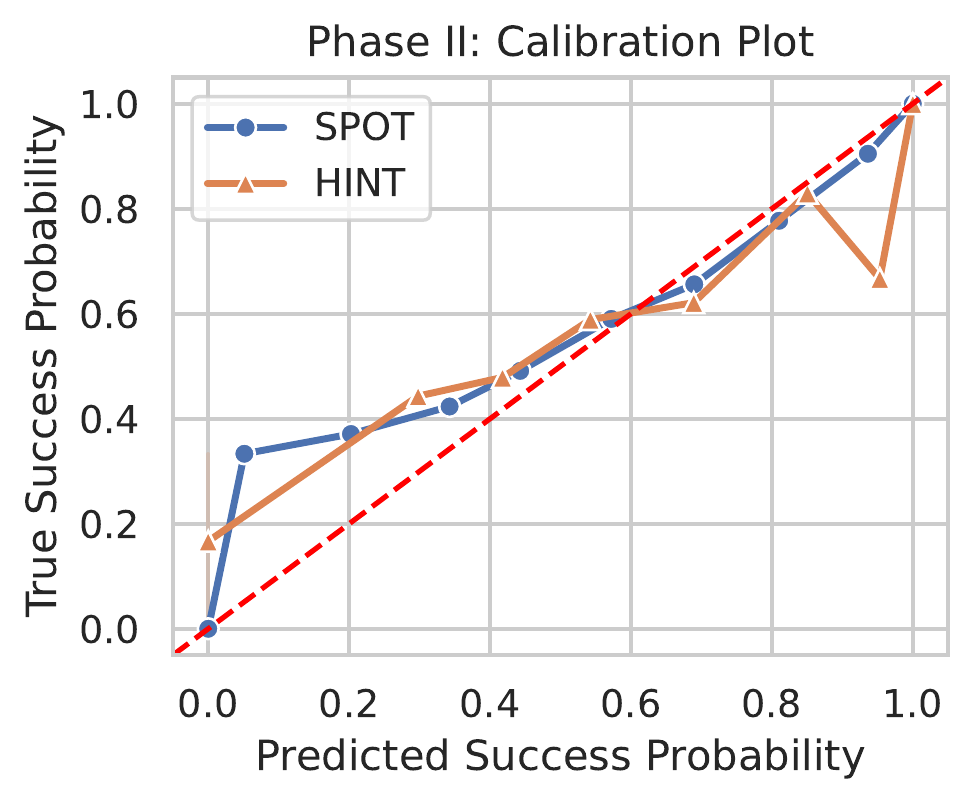}
\end{subfigure}
\hfill
\begin{subfigure}[b]{0.33\linewidth}
\includegraphics[width=\linewidth]{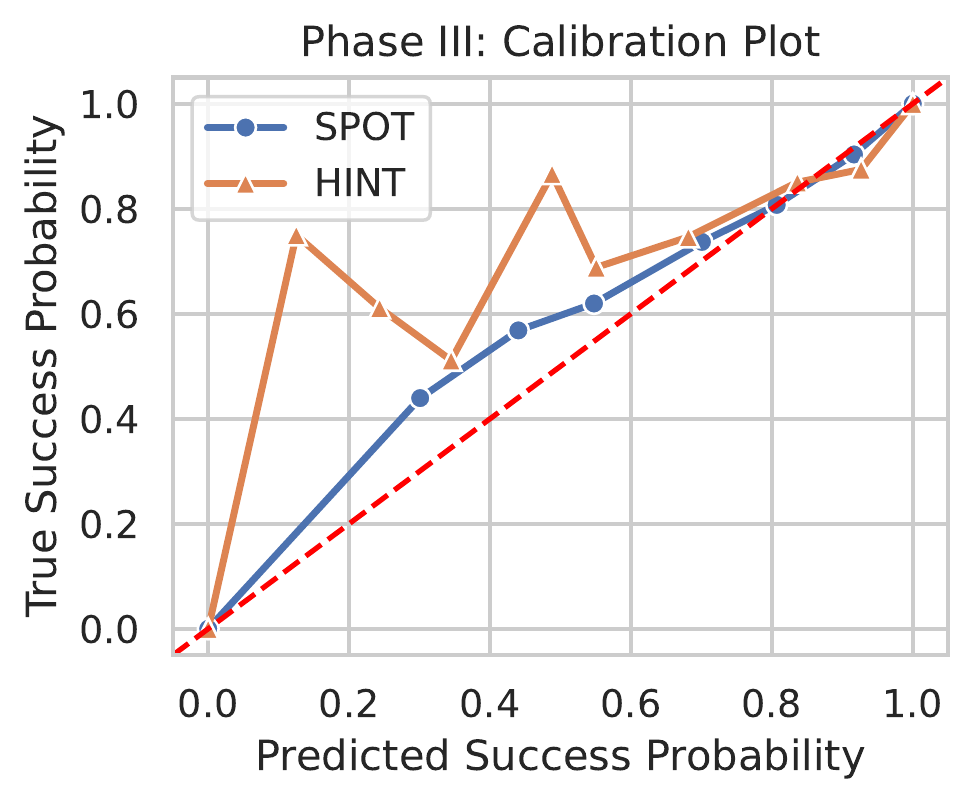}
\end{subfigure}
\caption{The calibration plots of our method \method and the best baseline HINT. We split the raw trials into a series of subgroups and calculate the average ground truth and predicted success rate of these groups. The red dotted line is the theoretical upper bound of the performance where the prediction is perfectly calibrated, such that the closer to it the better the prediction.}\label{fig:calibration}
% \vspace{-1em}
\end{figure*}

\begin{figure}[t]
\centering
\begin{subfigure}[b]{0.66\linewidth}
\includegraphics[width=\linewidth]{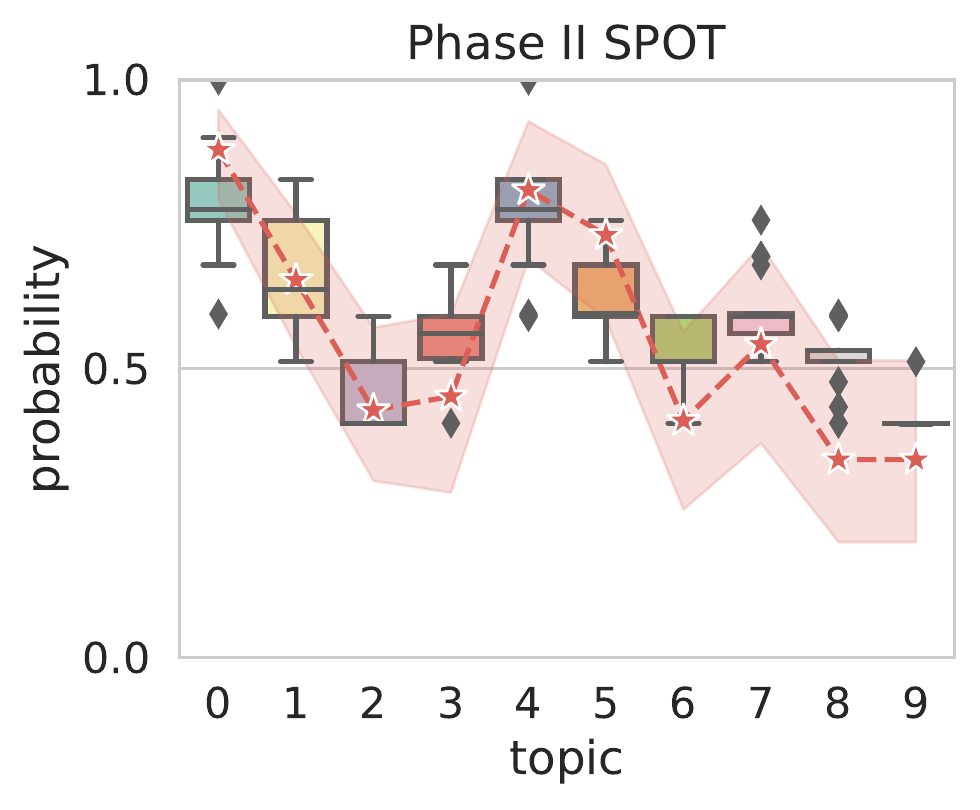}
\end{subfigure}
\hfill
\begin{subfigure}[b]{0.66\linewidth}
\includegraphics[width=\linewidth]{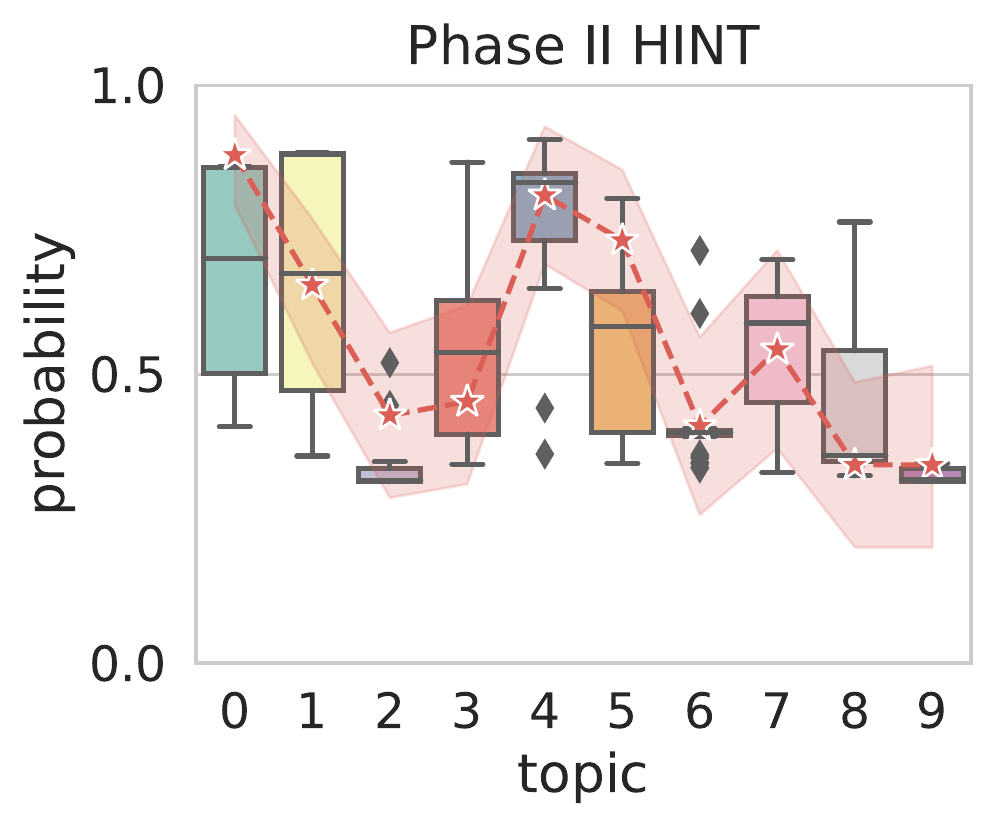}
\end{subfigure}
\caption{Predicted probability distributions of \method and HINT. The x-axis shows topic indices determined by \method. The red line indicates the actual average success rates of different topics with associated confidence intervals. \label{fig:disparity}}
\end{figure}

% {\color{red}calibration scatter plots}\cx{Is this a note?}

In addition, from Tables \ref{tab:result_phase_I}, \ref{tab:result_phase_II}, \ref{tab:result_phase_III}, we also observe the superior performance of \method in terms of PR-AUC: it obtains 8.9\% lift in phase II trials and 5.5\% lift in phase III trials, respectively, which indicates that \method has the strength in making well-calibrated probability predictions. Nonetheless, it is found that \method is slightly weaker than HINT in ROC-AUC for these trials, which implies it's a potential drawback of ranking the success probabilities across trials. We dive deep to investigate the cause, plotting the comparison between the probability distributions of \method and HINT in Fig. \ref{fig:disparity}. In particular, we divide the trials by their topics that are assigned by the topic discovery module of \method. We also plot the ground truth average success rates of each topic in red dotted lines as a reference. 

Fig. \ref{fig:disparity} shows the predictions generated by \method exhibit a higher degree of concentration (i.e., narrow ranges across different topics) around the ground truth average success rates (red dash lines), compared to the predictions generated by HINT (with wider ranges across topics and not calibrated with ground truth). This confirmed the benefit of the task-specific modeling employed by \method, which utilizes a distinct set of parameters for each task. This feature enables \method to more accurately capture the actual success probabilities of trials, taking into account the specific trial topic. %This is especially useful for trials targeting rare diseases. 
 %However, it should be noted that this may lead to a decrease in performance in terms of ranking trials across different topics due to the probable overestimation of the success probability of the trials belonging to the topic with a high mean success rate.

%We argue it is essential to note that when planning a clinical trial, the primary focus for investigators should be on the specific details and likelihood of success for the trial under consideration, rather than comparing it to trials from distinct topics. This is because a precise understanding of the probability of success for the trial being planned can aid investigators in making informed decisions about whether to continue with the trial or terminate it early. Thus, \method's ability to accurately capture the actual success probabilities of trials, accounting for the specific topic of the trial, is of paramount importance in the planning and execution of clinical trials.

\begin{figure*}[h!]
\centering
\begin{subfigure}[b]{0.33\linewidth}
\includegraphics[width=\linewidth]{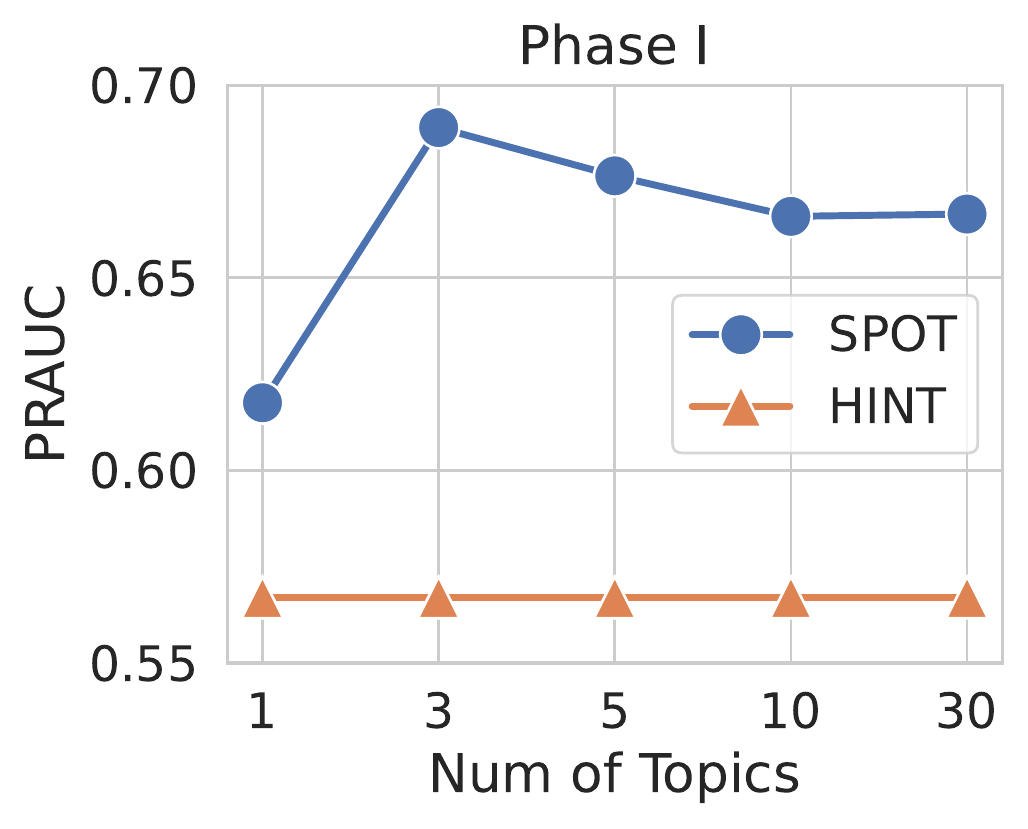}
\end{subfigure}
\hfill
\begin{subfigure}[b]{0.33\linewidth}
\includegraphics[width=\linewidth]{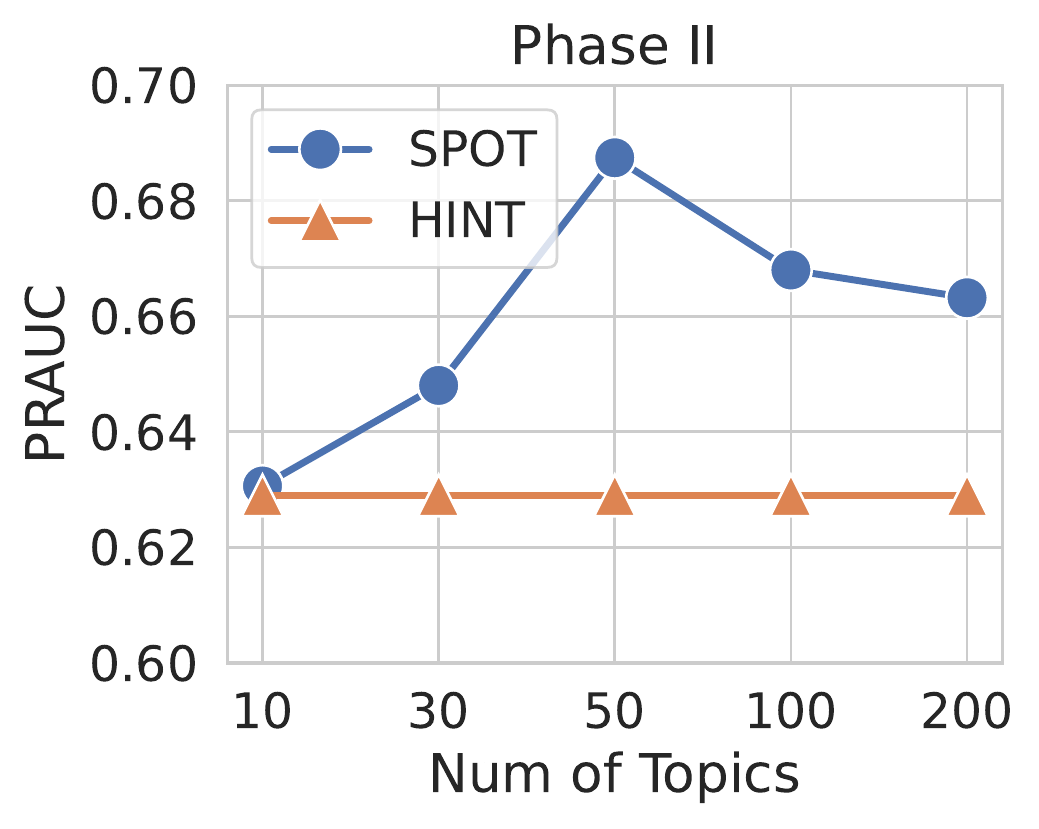}
\end{subfigure}
\hfill
\begin{subfigure}[b]{0.33\linewidth}
\includegraphics[width=\linewidth]{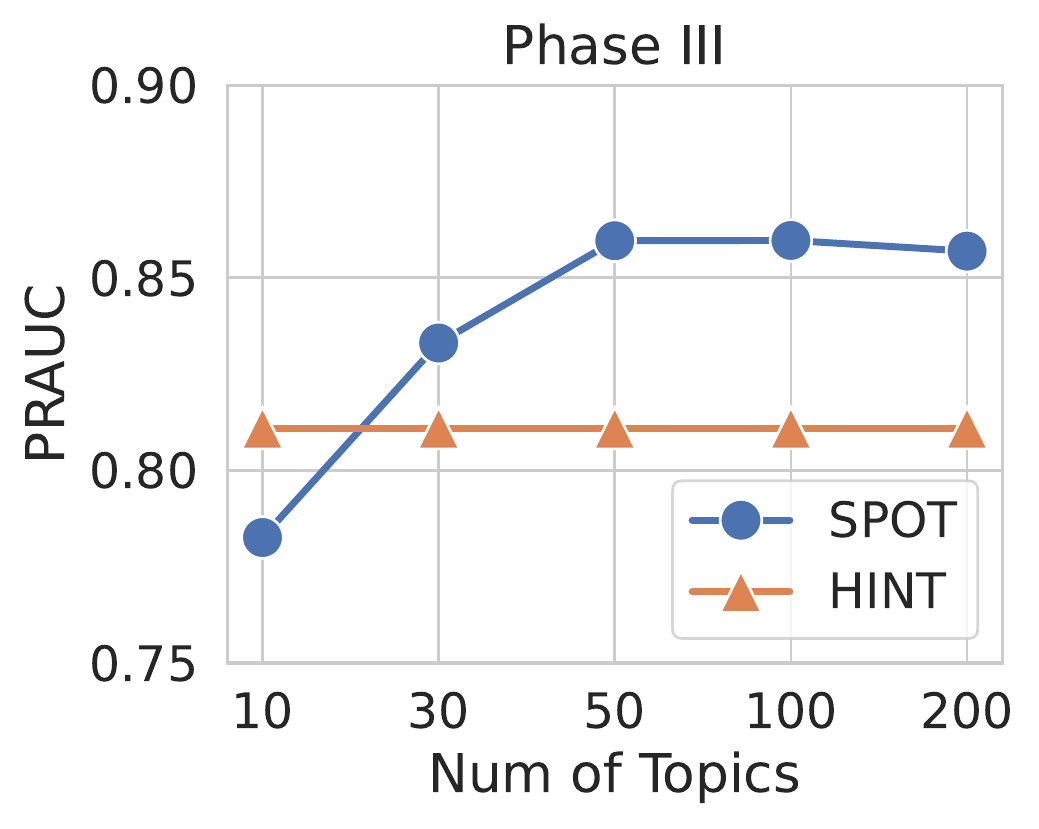}
\end{subfigure}
\caption{The analysis of \method on the performance varies according to the adjusted number of topics. \label{fig:sensitivity_analysis}}
\end{figure*}

\subsection*{Exp 2. Performance Across Trial Groups}\label{sec:exp_disparity}
To assess if the predicted trial outcomes are well calibrated with the true probabilities, we report the calibration plot of our method \method and HINT in Fig. \ref{fig:calibration}, for phases I, II, and III trial data, respectively. The calibration plot is often used in ML applications such as classification, where the aim is to predict the probability of an event, such as success or failure. In a well-calibrated model, the predicted probabilities should align with the actual outcomes, with high predicted probabilities indicating high actual probabilities, and low predicted probabilities indicating low actual probabilities. 

In Fig. \ref{fig:calibration}, it is witnessed that \method (blue line) is in general closer to the perfect calibration line (the 45-degree diagonal red dotted line), especially for phase II and III trials. That means, the \method's predicted probability can be a better estimate of the true trial success probability. Specifically, we find HINT is prone to overestimating the success probability for those success trials while underestimating the probability for those failure trials. For instance, in the phase I calibration plot (left in Fig. \ref{fig:calibration}), in the region where $x \in [0, 0.5]$, the HINT line is above the perfect line, meaning the predicted trials with around 35\% success rates have an actual success rate of around 50\%; while in the region $x \in [0.6, 0.8]$, the HINT line is below the perfect line, meaning the predicted trials with around 70\% success rates only have an actual success rate of around 58\%.

In comparison, \method typically aligns closely with the perfect line in the range $x \in [0.5, 1.0]$ for all three phases and delivers a more accurate prediction quality for the failure trials. For failure trials, \method tends to slightly overestimate in phase I while underestimating in phases II and III, reflecting the challenges in predicting these outcomes due to the unpredictable nature of complex factors that cause the failure of trials.

We also report the relative success and failure proportion divided by the predicted success probability on the test set in Fig. \ref{fig:relative_proportion} for further visualization of the predictions made by HINT and \method. To be specific, our method is more accurate in identifying failed trials, as they constitute a larger proportion of the group with the lowest predicted success probability. This feature can provide valuable insight for investigators to optimize their trial design and reduce the likelihood of executing failing trials. 
 On the other hand, HINT has been exhibited to be less accurate in identifying failed trials. This is attributed to the fact that failed trials do not constitute a majority of the group with the lowest predicted success probability by HINT. Additionally, the success-failure ratio across groups illustrates that the predictions made by \method deviate less from the ground truth ratio. It is worth noting that, generally, the identification of failed trials is a more challenging task than the identification of successful trials for both. Despite it, the superior performance of \method in identifying failed trials renders it a valuable tool for investigators to improve the design and execution of clinical trials.

\subsection*{Exp 3. In-depth Analysis}\label{sec:exp_sensitivity}

\textbf{Sensitivity Analysis.}
A key insight of \method is to build the topic-specific models because the trial outcomes often depend on their trial topics. The number of topics $K$ becomes one critical hyperparameter that needs to be properly chosen. To test the sensitivity of our method with respect to $K$, we train \method with varying $K$ on three phases of trial data separately. We set $K \in \{1, 3, 5, 10, 30\}$ for phase I trials, $K \in \{10, 30, 50, 100, 200\}$ for phase II and III trials, respectively. The results are reported in Fig. ~\ref{fig:sensitivity_analysis}. The ranges of $K$ for phase II and III are larger than that of phase I because there are more trials from phase II and III in the training data.

As shown in Fig~\ref{fig:sensitivity_analysis}, \method is reasonably robust to different topic numbers $K$ as \method almost always outperforms HINT with all $K$ selections. This is particularly useful as the optimal number of topics can vary depending on the size and complexity of the data. In particular, the optimal topic number for phase I trials is 3 while is 50 for phases II and III in our experiments. The training data of phase II and III trials are both around three times bigger than phase I and the primary measures for the later stage trials are much more diverse than the early ones. Most phase I trials focus on evaluating the treatment safety while the phase II and III trials will also consider the efficacy considering distinct diseases. Moreover, phases II and III trials typically involve a larger number of patients and a wider range of diseases, which can result in a more diverse set of outcomes. This increased diversity may require a more granular topic-aware modeling approach to capture the nuances of the data.\\

\begin{figure*}[h!]
\centering
\begin{subfigure}[b]{0.33\linewidth}
\includegraphics[width=\linewidth]{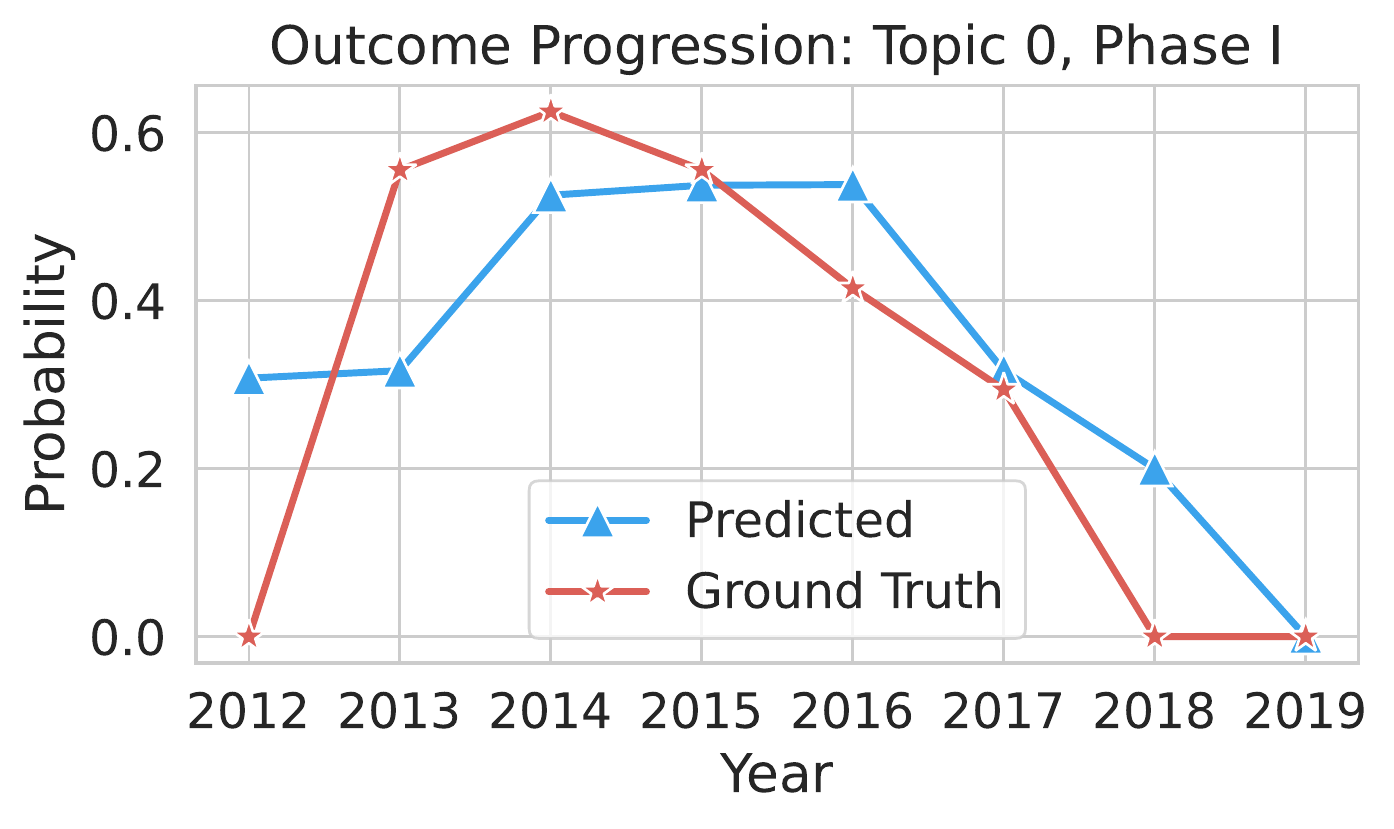}
\end{subfigure}
\hfill
\begin{subfigure}[b]{0.33\linewidth}
\includegraphics[width=\linewidth]{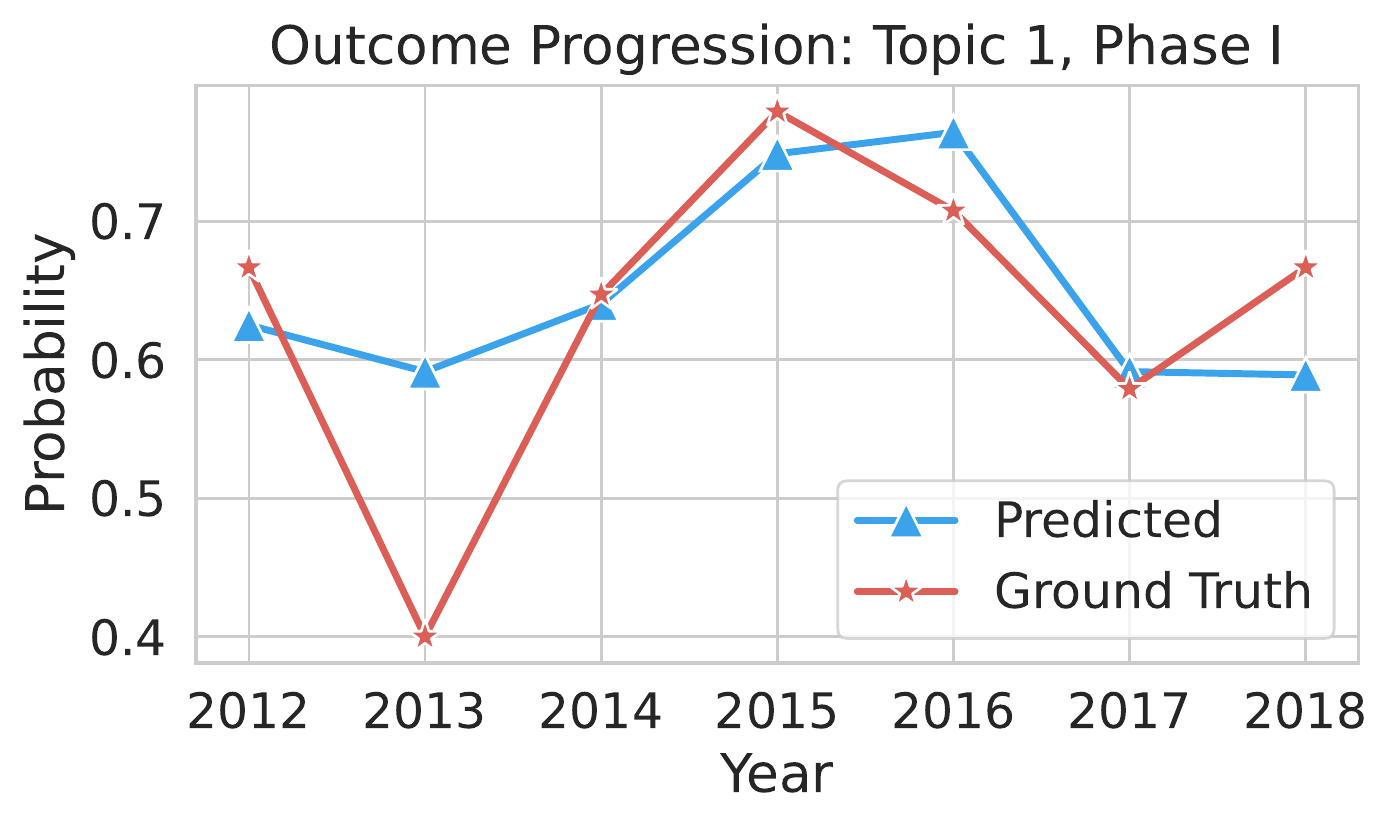}
\end{subfigure}
\hfill
\begin{subfigure}[b]{0.33\linewidth}
\includegraphics[width=\linewidth]{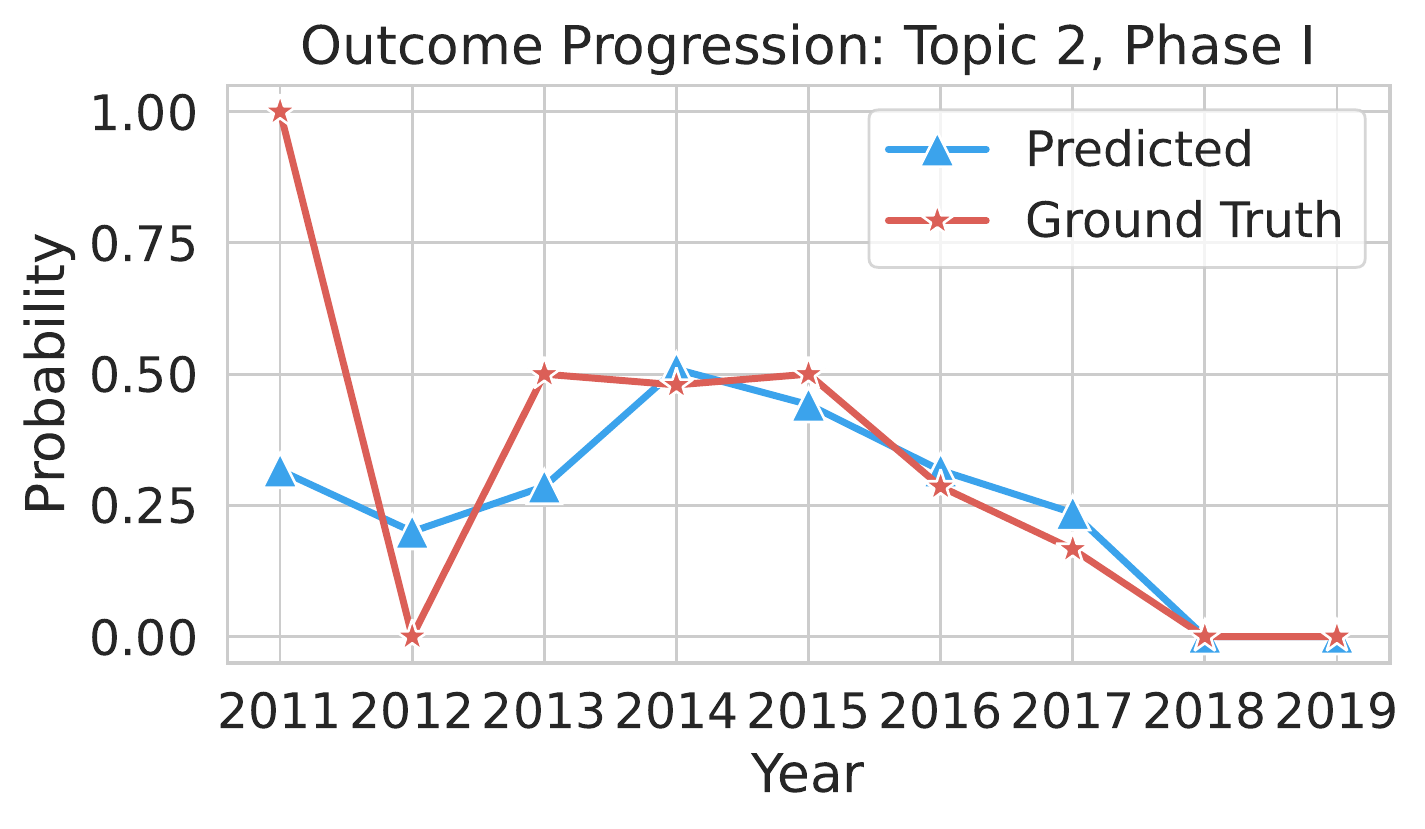}
\end{subfigure}
\caption{The analysis of \method on modeling the progression of clinical trial outcomes along the timesteps within each topic. The results are obtained on the test set of phase I trials. 
\label{fig:progression}}
\end{figure*}

\noindent \textbf{Trial Outcome Progression.} The ability of \method to model the sequential progression of trial outcomes within each trial topic is a key feature that sets it apart from other methods. This is because it allows for a deeper understanding of the temporal patterns of trial outcomes and how they are affected by technological advancement. In Fig. \ref{fig:progression}, we present several examples of the progression modeling results in the topics by our method. These examples are specifically chosen from the test set of phase I trials, and they show the average ground truth success rates of the trials, as well as the average predicted probabilities, along the timestep in the topic.

The result illustrates that \method is able to accurately capture the temporal progression of trial outcomes as the predicted probabilities closely match the ground truth success rates. The ability to track the success rate over time allows an understanding of how the technology and the treatments evolve. It can also identify the points where the success rate is increasing or decreasing and by how much. This can provide valuable insights for researchers to track the progress of drug development and identify the trends over time, which can inform future research and development.

\subsection*{Exp 4. Ablation Study} \label{sec:exp_ablation}
Our method constitutes three main components including the topic discovery module, the sequential modeling module, and the meta-learning module. We conduct ablations on them to estimate the effects of each component contributed to the final performance. We test the variants that remove topic discovery (\texttt{w/o Topic}), and remove sequential modeling (\texttt{w/o Sequence}), and remove meta-learning (\texttt{w/o Meta-Learning}). Note that the \texttt{w/o Topic} variant also does not include meta-learning because we only have one task in this case. We report the results in Table \ref{tab:ablation}. 

The results reveal that the topic discovery module has a significant effect on the overall performance. Specifically, when evaluating the results of phase II and III trials, it is evident that the \texttt{w/o Topic} variant performs significantly worse, with a low ROC-AUC score. The reason for this poor performance is that aggregating all trials together and ordering them in chronological order creates a random temporal pattern, resulting in trivial predictions. On the other hand, incorporating sequential modeling improves the performance of phase I trials significantly. But its impact is less significant in Phase II and III trials, where the temporal pattern is more complex.

\begin{table}[t]
  \centering
  \caption{Ablation analysis of \method when each module is removed. w/o Topic: without topic discovery; w/o Sequence: without sequential modeling; w/o Meta-Learning: without meta-learning.}
         \resizebox{0.8\linewidth}{!}{%
    \begin{tabular}{lccc}
    \toprule
     \textbf{Method} & \multicolumn{1}{c}{\textbf{PR-AUC}} & \multicolumn{1}{c}{\textbf{F1}} & \multicolumn{1}{c}{\textbf{ROC-AUC}} \bigstrut[b]\\ \hline
    \textbf{Phase I Trials} &       &       &  \bigstrut[t]\\
    SPOT  & 0.689 & 0.714 & 0.660 \bigstrut[t]\\
    w/o Topic & 0.604 & 0.713 & 0.569 \\
    w/o Sequence & 0.657 & 0.713 & 0.630 \\
    w/o Meta-Learning & 0.641 & 0.710 & 0.582 \bigstrut[b]\\
    \midrule
    \textbf{Phase II Trials} &       &       &  \bigstrut[t]\\
    SPOT  & 0.685 & 0.656 & 0.630 \bigstrut[t]\\
    w/o Topic & 0.573 & 0.521 & 0.516 \\
    w/o Sequence & 0.678 & 0.649 & 0.621 \\
    w/o Meta-Learning & 0.663 & 0.646 & 0.615 \bigstrut[b]\\
    \midrule
    \textbf{Phase III Trials} &       &       &  \bigstrut[t]\\

    SPOT  & 0.856 & 0.857 & 0.711 \bigstrut[t]\\
    w/o Topic & 0.769 & 0.813 & 0.518 \\
    w/o Sequence & 0.850 & 0.857 & 0.699 \\
    w/o Meta-Learning & 0.799 & 0.812 & 0.685 \bigstrut[b]\\
    \bottomrule
    \end{tabular}%
    }
  \label{tab:ablation}%
\end{table}%

\section{Conclusion}\label{sec:conclusion}
In conclusion, the accurate prediction of clinical trial outcomes is crucial for saving time and funding for pharmaceutical companies, as it allows for the early detection of trial failure. Previous approaches, however, suffer from poor performance for trials belonging to less common topics. In this work, we propose a meta-learning approach, called \method, which models a \textit{topic} of clinical trials in sequence instead of modeling each trial independently. By using trial topic discovery to define the topic, our approach can take into account the relations among trials and the distinction between groups of trials. Our results show that \method outperforms previous methods by a significant margin.

Future work for this approach could include incorporating additional sources of information, such as patient data or genetic information, to improve the performance of the model. More research could also be done on the interpretability of the model, such as identifying which factors have the most impact on trial outcomes. Additionally, it may be of interest to develop more sophisticated topic discovery method thus enhancing the task-aware modeling for clinical trial outcome predictions.

\bibliographystyle{ACM-Reference-Format}
\bibliography{main}

\appendix

% Table generated by Excel2LaTeX from sheet 'Sheet2'
\begin{table*}[t]
  \centering
  \caption{The dataset statistics of the trial's target disease groups, in phase I, II, and III clinical trials.}
         \resizebox{\linewidth}{!}{%
    \begin{tabular}{p{13em}c|p{13em}c|p{13em}c}
    \hline
    \multicolumn{2}{p{13em}|}{\textbf{Phase I}} & \multicolumn{2}{p{13em}|}{\textbf{Phase II}} & \multicolumn{2}{p{13em}}{\textbf{Phase III}} \bigstrut[t]\\
    Disease group & \multicolumn{1}{p{4.215em}|}{Num} & Disease group & \multicolumn{1}{p{4.215em}|}{Num} & Disease group & \multicolumn{1}{p{4.215em}}{Num} \bigstrut[b]\\
    \hline
    Neoplasms                                                                                               & 552   & Neoplasms                                                                                               & 1931  & Neoplasms                                                                                               & 637 \bigstrut[t]\\
    Factors Influencing Health Status and Contact with Health Services                                      & 149   & Factors Influencing Health Status and Contact with Health Services                                      & 405   & Endocrine, Nutritional and Metabolic Diseases                                                           & 447 \\
    Endocrine, Nutritional and Metabolic Diseases                                                           & 99    & Injury, Poisoning and Certain Other Consequences of External Causes                                     & 288   & Diseases of the Nervous System                                                                          & 392 \\
    Diseases of the Nervous System                                                                          & 71    & Diseases of the Digestive System                                                                        & 284   & Diseases of the Circulatory System                                                                      & 361 \\
    Certain Infectious and Parasitic Diseases                                                               & 70    & Diseases of the Nervous System                                                                          & 276   & Factors Influencing Health Status and Contact with Health Services                                      & 351 \\
    Injury, Poisoning and Certain Other Consequences of External Causes                                     & 70    & Diseases of the Circulatory System                                                                      & 267   & Mental, Behavioral and Neurodevelopmental Disorders                                                     & 329 \\
    Diseases of the Eye and Adnexa                                                                          & 69    & Endocrine, Nutritional and Metabolic Diseases                                                           & 258   & Certain Infectious and Parasitic Diseases                                                               & 290 \\
    Diseases of the Digestive System                                                                        & 54    & Certain Infectious and Parasitic Diseases                                                               & 254   & Diseases of the Digestive System                                                                        & 284 \\
    Diseases of the Circulatory System                                                                      & 52    & Diseases of the Respiratory System                                                                      & 242   & Diseases of the Respiratory System                                                                      & 272 \\
    Diseases of the Genitourinary System                                                                    & 47    & Diseases of the Genitourinary System                                                                    & 230   & Diseases of the Genitourinary System                                                                    & 271 \\
    Certain Conditions Originating in the Perinatal Period                                                  & 32    & Diseases of the Musculoskeletal System and Connective Tissue                                            & 214   & Certain Conditions Originating in the Perinatal Period                                                  & 247 \\
    Symptoms, Signs and Abnormal Clinical and Laboratory Findings, Not Elsewhere Classified                 & 28    & Mental, Behavioral and Neurodevelopmental Disorders                                                     & 194   & Diseases of the Musculoskeletal System and Connective Tissue                                            & 227 \\
    Diseases of the Musculoskeletal System and Connective Tissue                                            & 28    & Certain Conditions Originating in the Perinatal Period                                                  & 133   & Diseases of the Eye and Adnexa                                                                          & 189 \\
    Diseases of the Respiratory System                                                                      & 27    & Diseases of the Eye and Adnexa                                                                          & 103   & Injury, Poisoning and Certain Other Consequences of External Causes                                     & 114 \\
    Mental, Behavioral and Neurodevelopmental Disorders                                                     & 21    & Congenital Malformations, Deformations and Chromosomal Abnormalities                                    & 96    & Symptoms, Signs and Abnormal Clinical and Laboratory Findings, Not Elsewhere Classified                 & 88 \\
    Congenital Malformations, Deformations and Chromosomal Abnormalities                                    & 19    & Diseases of the Blood and Blood Forming Organs and Certain Disorders Involving the Immune Mechanism     & 83    & Diseases of the Blood and Blood Forming Organs and Certain Disorders Involving the Immune Mechanism     & 84 \\
    Diseases of the Blood and Blood Forming Organs and Certain Disorders Involving the Immune Mechanism     & 18    & Diseases of the Skin and Subcutaneous Tissue                                                            & 80    & Congenital Malformations, Deformations and Chromosomal Abnormalities                                    & 77 \\
    Diseases of the Skin and Subcutaneous Tissue                                                            & 11    & Symptoms, Signs and Abnormal Clinical and Laboratory Findings, Not Elsewhere Classified                 & 51    & Pregnancy, Childbirth and the Puerperium                                                                & 44 \\
    External Causes of Morbidity                                                                            & 11    & Pregnancy, Childbirth and the Puerperium                                                                & 25    & Diseases of the Skin and Subcutaneous Tissue                                                            & 40 \\
    Pregnancy, Childbirth and the Puerperium                                                                & 9     & External Causes of Morbidity                                                                            & 22    & External Causes of Morbidity                                                                            & 19 \\
    Diseases of the Ear and Mastoid Process                                                                 & 5     & Diseases of the Ear and Mastoid Process                                                                 & 5     & Diseases of the Ear and Mastoid Process                                                                 & 5 \bigstrut[b]\\
    \hline
    \end{tabular}%
    }
  \label{appx:tab:disease_group_stats}%
\end{table*}%

\begin{figure*}[t]
\centering
\begin{subfigure}[b]{0.33\linewidth}
\includegraphics[width=\linewidth]{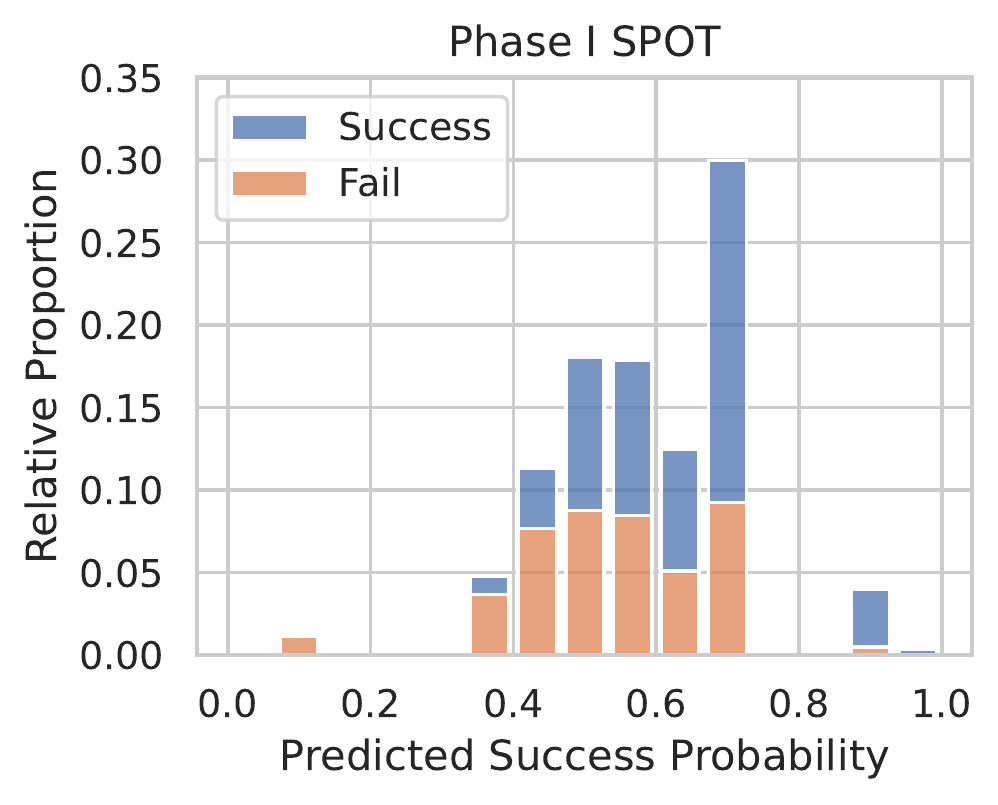}
\end{subfigure}
\hfill
\begin{subfigure}[b]{0.33\linewidth}
\includegraphics[width=\linewidth]{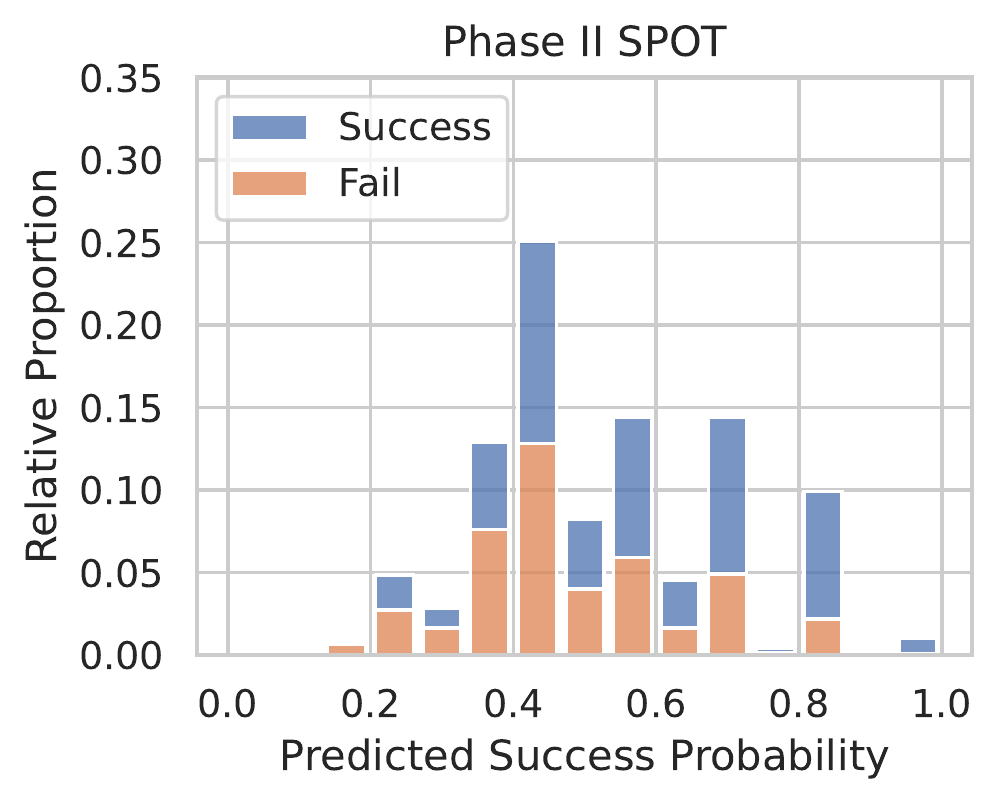}
\end{subfigure}
\hfill
\begin{subfigure}[b]{0.33\linewidth}
\includegraphics[width=\linewidth]{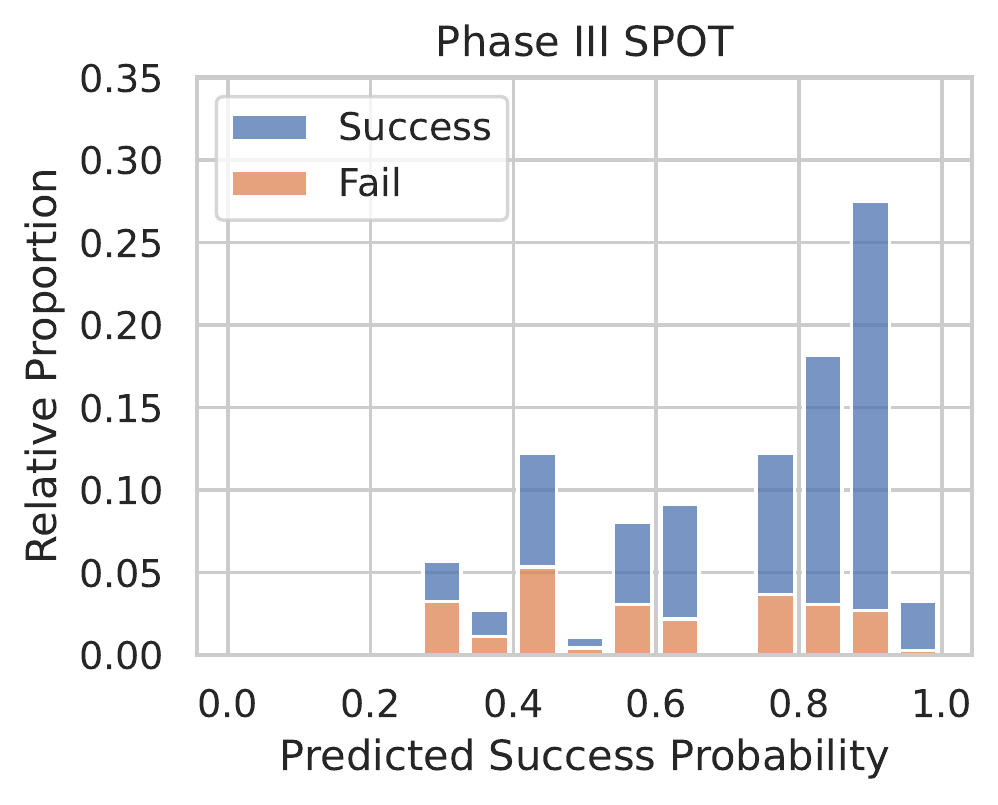}
\end{subfigure}
\hfill
\begin{subfigure}[b]{0.33\linewidth}
\includegraphics[width=\linewidth]{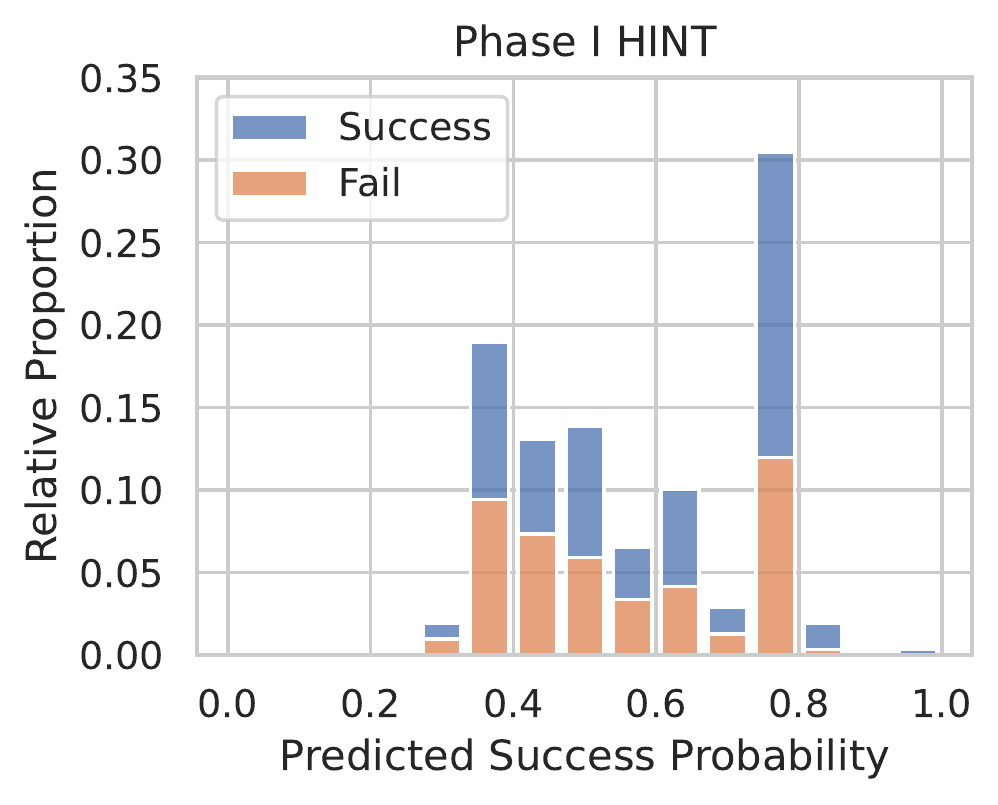}
\end{subfigure}
\hfill
\begin{subfigure}[b]{0.33\linewidth}
\includegraphics[width=\linewidth]{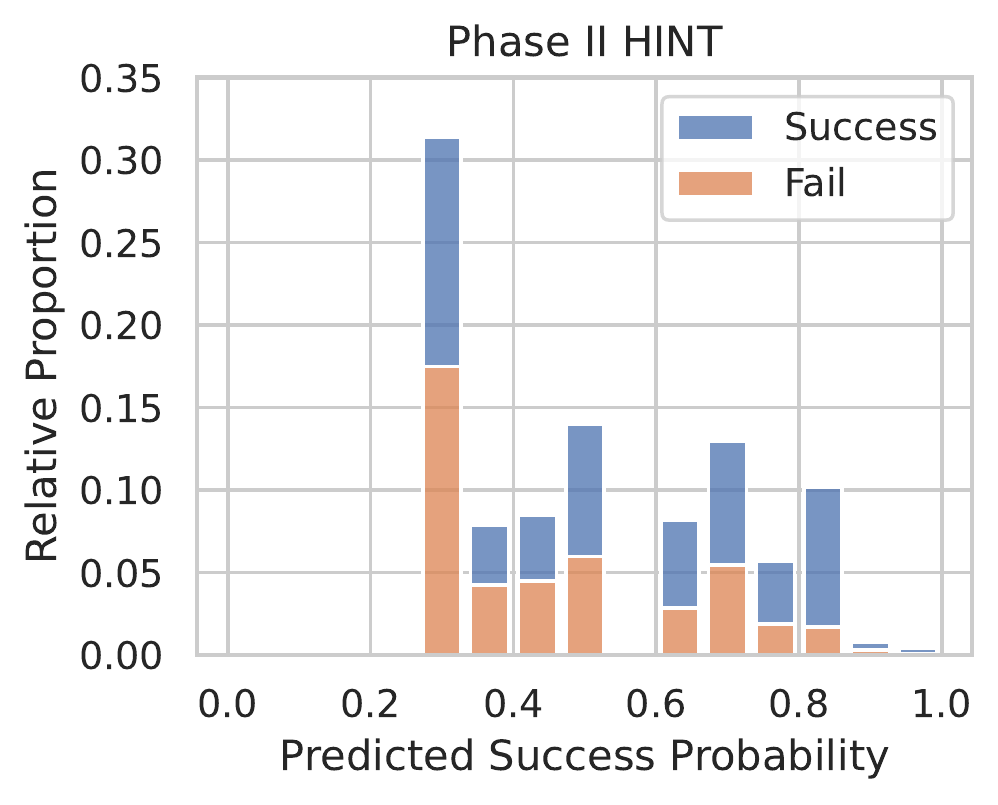}
\end{subfigure}
\hfill
\begin{subfigure}[b]{0.33\linewidth}
\includegraphics[width=\linewidth]{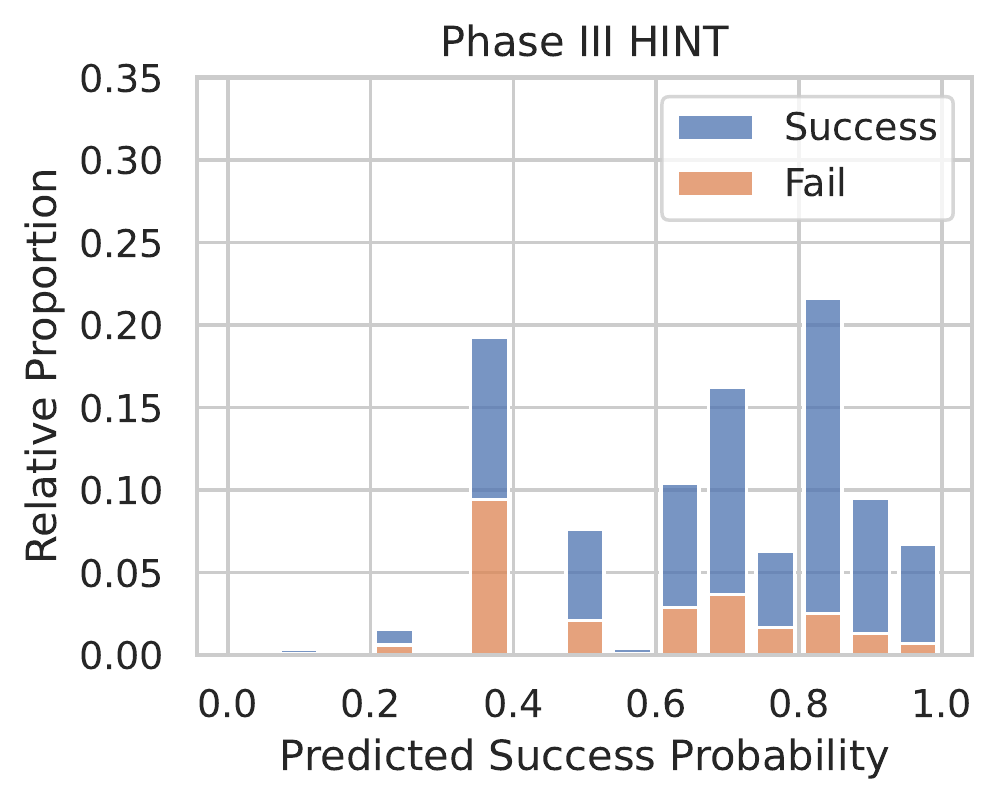}
\end{subfigure}
\caption{The predicted success probability versus relative success/failure proportion of our method \method (top) and the baseline HINT (bottom). The predicted probabilities of \method reflect the actual distribution of success and failure trials, where failure trials are mostly found in the low predicted probability group. On the other hand, the baseline method HINT shows a less distinct division between success and failure trials in its predicted probabilities.}\label{fig:relative_proportion}
\vspace{-1em}
\end{figure*} 

\end{document}